\documentclass{article} 
\usepackage{iclr2027_conference,times}

\usepackage{amsmath,amsfonts,bm}

\def\eqref#1{equation~\ref{#1}}

\def\1{\bm{1}}

\DeclareMathAlphabet{\mathsfit}{\encodingdefault}{\sfdefault}{m}{sl}
\SetMathAlphabet{\mathsfit}{bold}{\encodingdefault}{\sfdefault}{bx}{n}

\usepackage{hyperref}
\usepackage{url}
\usepackage{booktabs}
\usepackage{nicefrac}
\usepackage{microtype}
\usepackage{wrapfig}
\usepackage{soul}
\usepackage{tikz,adjustbox}
\usepackage[capitalize,noabbrev]{cleveref}
\usepackage{fancyvrb}

\usepackage{inconsolata}
\usepackage[skins,breakable,most]{tcolorbox}

\usepackage{caption}
\usepackage{subcaption}
\usepackage{enumitem}
\usepackage{flushend}
\usepackage{balance}
\usepackage{lineno}
\usepackage{amsmath,amssymb,amsfonts}
\usepackage{algorithm}
\usepackage{algpseudocode}
\usepackage{graphicx}
\usepackage{textcomp}
\usepackage[table]{xcolor}
\usepackage{xspace}
\usepackage{colortbl}
\usepackage{amsthm}
\usepackage{float}
\usepackage{multirow}
\usepackage{multicol}
\usepackage{bm}
\usepackage{bbm}
\usepackage{placeins}
\usepackage{cuted}
\usepackage{fontawesome5}

\usepackage{titletoc}
\newcommand\DoToC{
  \startcontents
  \printcontents{}{1}{\textbf{Contents of Appendix}\vskip3pt\hrule\vskip5pt}
  \vskip3pt\hrule\vskip5pt
}

\title{ExpVoyager: Direct Experience Navigation for Dynamic Agent Skill Synthesis}

\newif\ifarxivversion
\arxivversiontrue

\usepackage{etoolbox}
\usepackage{pifont}

\definecolor{linkpink}{HTML}{C2185B}

\hypersetup{
  pdftitle={ExpVoyager: Direct Experience Navigation for Dynamic Agent Skill Synthesis}
}

\ifarxivversion

  \iclrfinalcopy

  \author{%
    \textbf{Kwangwook Seo}\qquad
    \textbf{Dongha Lee}\thanks{Corresponding author} \\
    \raisebox{-0.12em}{%
      \includegraphics[height=0.95em]{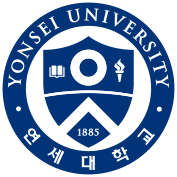}%
    }\hspace{0.35em}Yonsei University \\
    \texttt{\{tommy2130,donalee\}@yonsei.ac.kr}
  }

  \hypersetup{
    colorlinks=true,
    citecolor=linkpink,
    linkcolor=linkpink,
    urlcolor=linkpink,
    pdfauthor={Kwangwook Seo and Dongha Lee}
  }

  \renewcommand{\headrulewidth}{0.4pt}

  \makeatletter

  \pretocmd{\@maketitle}
    {}
    {}
    {\PackageError{expvoyager}{Corresponding author symbol patch failed}
      {Check the conference style.}}

  \patchcmd{\@maketitle}
    {\vskip 0.3in minus 0.1in}
    {\vskip 0.05in}
    {}
    {\PackageError{expvoyager}{Title spacing patch failed}
      {Check the conference style.}}

  \apptocmd{\@maketitle}
    {\fancyhead{}\fancyhead[L]{\small\textnormal{Preprint}}}
    {}
    {\PackageError{expvoyager}{Preprint header patch failed}
      {Check the conference style.}}

  \makeatother

\else

  \iclrfinalfalse

  \hypersetup{
    colorlinks=false,
    pdfauthor={}
  }

  \renewcommand{\faGithub}{}

\fi

\begin{document}

\maketitle

\begin{abstract}
Learning from experience in LLM agents has become a key paradigm for developing self-evolving agents that continuously learn and expand their capabilities.
Within this paradigm, synthesizing the \textit{agent skill} has emerged as a promising solution for transforming accumulated experience into reusable \textit{procedural knowledge}, serving as an important layer for the harness system that supplies agents at runtime.
Despite its potential, existing approaches largely abstract past experience into fixed procedural knowledge before downstream demands are known, which risks discarding knowledge that later becomes critical while retaining instance-specific details irrelevant to future tasks.
In this paper, we reframe agent skill synthesis as a dynamic navigation problem over past experience, where agents actively explore accumulated trajectories on demand for the current task with targeted and fine-grained access to experience knowledge.
To this end, we propose \textbf{\textsc{ExpVoyager}}, a novel framework in which a skill curator navigates raw experience across different views and resolutions, continually identifying reusable procedural knowledge from what it observes while tracking remaining knowledge needs that guide where to navigate next.
Extensive experiments demonstrate both the effectiveness and versatility of \textsc{ExpVoyager}, showing consistent improvements in downstream task performance, continual gains as the experience space scales, and practical compatibility with existing skills under efficient experience access. 
\href{https://github.com/tommyEzreal/ExpVoyager}{\faGithub\ \textbf{\texttt{[CODE]}}}
\end{abstract}
\section{Introduction}
LLMs are rapidly evolving from systems that merely generate responses into long-horizon agents that interact with environments, use tools, and execute complex multi-step tasks~\citep{yao2023react,pmlr-v205-ichter23a}. 
As the tasks and environments they operate in become increasingly diverse and complex, the general capabilities of foundation models alone may not be sufficient to reliably provide the knowledge required across different execution contexts, making the \textit{agent harness}~\citep{openai2026harness,lee2026metaharness,zhang2026selfharnessharnessesimprove} an increasingly important layer for supplying agents at runtime.

In response to these needs, the \textit{agent skill}~\citep{anthropic2025agentskills}, typically instantiated as manually authored instructions, procedures, and heuristics that guide agent behavior, has emerged as a promising solution for extending agent capabilities without modifying model weights.
Recently, a line of research~\citep{wang2024voyager,wang2025inducing} has taken a step further by enabling agents to synthesize skills from their own task experience, showing the potential of self-evolving agents~\citep{10.1609/aaai.v38i17.29936,zheng2025skillweaverwebagentsselfimprove,ouyang2026reasoningbank} that continuously learn and expand their capabilities.

One of the key challenges in this direction is to transform past execution experience from task-specific records into \textit{procedural knowledge}~\citep{wu2026proceduralknowledgescaleimproves,mi2026skillpro} that generalizes to future decision-making.
Existing approaches largely address this challenge by abstracting reusable lessons from past experience into fixed procedural knowledge, either summarizing individual trajectory~\citep{ouyang2026reasoningbank,fang-etal-2026-memp} or consolidating patterns across multiple trajectories~\citep{wang2025agent,ni2026trace2skilldistilltrajectorylocallessons}, and storing the resulting skills for future use.
However, as the demands of future tasks are inherently unknown at the time of skill construction, this approach risks discarding knowledge that later becomes critical, while retaining excessive instance-specific details that are irrelevant to their downstream application.
This gap leads us to ask a central question: \textbf{\textit{how can we enable agents to dynamically synthesize skills on demand for the current task from raw experience?}}

In this paper, we answer this question by reframing skill synthesis from the fixed abstraction of past experience into a \textbf{dynamic navigation problem}, where agents search for procedural knowledge on demand within accumulated trajectories.
One straightforward instantiation of this task-time formulation is to retrieve the top-\textit{k} trajectories most relevant to the current task and synthesize a task-specific skill from them.
However, similarity-based access to past experience can struggle to capture procedural relevance to the current task, as irrelevant knowledge is often contained in similar trajectories, while localized critical cues may also appear in superficially dissimilar trajectories.
Therefore, instead of relying on such a passive retrieval interface, we define experience navigation as an active decision-making process carried out by the agent itself, in which the agent determines where to inspect and at what resolution to examine past experience targeted to the current needs.

To this end, we introduce \textbf{\textsc{ExpVoyager}}, a novel framework for dynamic agent skill synthesis in which a skill curator navigates past experience to construct task-specific guidance for a frozen executor.
Motivated by the intuition that useful procedural knowledge needed for a target task often resides in different aspects of a trajectory (such as overall execution patterns and local decision contexts), we first equip the curator with a \textit{\textbf{Navigable Interface}} that supports access to multiple views of raw experience at different resolutions. 
Through this interface, the curator actively decides how to access past experience by selecting navigation actions that specify the appropriate level of view in its action arguments for the current investigation, and expand localized records into broader trajectory context when needed.
Next, to support the navigation process that evolves with what the curator has observed in previously accessed experience, we also design a \textit{\textbf{Navigation State}} for the curator.
This state is updated each round by interpreting newly accessed observations in order to decide what procedural knowledge can be reused for the current target and what remains to be investigated, thereby linking experience access and interpretations to continually guide one another.

We conduct extensive experiments on three popular agent benchmarks including household interaction, online shopping, and scientific reasoning.
Our evaluation mainly focuses on (1) the \textit{effectiveness} of \textsc{ExpVoyager} in improving agent task performance and (2) its \textit{versatility} in supporting scalable and efficient experience reuse.
Across all benchmarks, \textsc{ExpVoyager} consistently outperforms existing skill-based approaches in task performance and improves base agents across diverse curator–executor configurations.
\textsc{ExpVoyager} also drives online self-evolution without relying on pre-collected experience, and additional experimens shows that it can scale to larger experience space, progressively widening its performance advantage over existing approaches.
We further demonstrate its versatility in practical agent-harness scenarios, showing that \textsc{ExpVoyager} can synergize with existing skills through on-demand refinement while also making effective use of experience-access budgets.

We summarize our contributions as follows:
\begin{itemize}[leftmargin=1.5em, labelsep=0.5em, itemsep=3pt, topsep=3pt]
    \item We reframe agent skill synthesis as a dynamic navigation problem over past experience, enabling targeted and fine-grained access to task-relevant procedural knowledge on demand.

    \item  We propose \textsc{ExpVoyager}, a novel framework that supports effective experience navigation with a \textit{Navigable Interface} for accessing raw experience through different views and resolutions, and \textit{Navigation State} that continually connects experience access with target-relevant knowledge.

    \item We demonstrate both the effectiveness and versatility of \textsc{ExpVoyager}, showing consistent improvements in downstream task performance, continual gains as the experience space scales, and practical compatibility with existing skills under efficient experience access.
\end{itemize}
\section{Preliminary Analysis}
\label{section2}

We conduct two preliminary analyses to better understand the limitations of existing skill construction and retrieval paradigms, asking: (1) how much knowledge useful for future tasks is preserved when skills are constructed in advance from past experience, and (2) how effectively existing retrieval paradigms can surface the knowledge needed for a given target task from past experience.

To empirically examine these questions, we first construct \textit{oracle skills} whose helpfulness for the target task is verified through task execution and annotate the \textit{oracle knowledge} in each source trajectory that can contribute to constructing each skill. Specifically, we first collect multiple successful and failed trajectories from distinct attempts of a frozen executor on each target task and use them to synthesize a candidate skill. We then provide the candidate skill back to the same frozen executor and retain it only when the executor successfully completes the corresponding task with the skill. After this iterative generate-then-verify process, we collect 80 oracle skills with verified helpfulness on ALFWorld~\citep{shridhar2021alfworld} test tasks. Next, to identify the oracle knowledge available in past experience for constructing each oracle skill, we sample 300 source trajectories from the ALFWorld training set and examine whether each source contains relevant knowledge. Helpful sources are annotated with the relevant oracle knowledge items, while the remaining sources are labeled \textit{none}. Please refer to Appendix~\ref{Preliminary} for more detailed experimental setups.

\setlength{\intextsep}{0pt}
\begin{wrapfigure}{r}{0pt}
\centering
\includegraphics[scale=0.9]{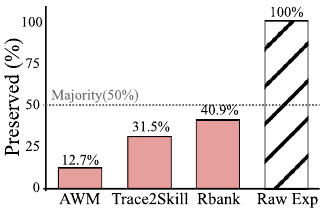}
\caption{Experiments on how much knowledge useful for future tasks is preserved in skills. Here, we use raw experience as the 100\% reference for preservation.}
\label{fig:analysis_1}
\end{wrapfigure}
\subsection{Preliminary Analysis I: Pre-Constructed Skills Lose the Majority of Procedural Knowledge Useful for Future Tasks.}
\paragraph{Setups.} 
We first examine how much knowledge that later becomes useful for a future task is preserved when past experience is abstracted into a skill before its downstream demand is known. Specifically, we compare the knowledge retained in pre-constructed skills of existing methods~\citep{wang2025agent,ouyang2026reasoningbank,ni2026trace2skilldistilltrajectorylocallessons} against the oracle knowledge from the corresponding source trajectory.
For quantification, we decompose each skill into atomic knowledge claims for semantic matching, following prior claim-level evaluation protocols~\citep{ru2024ragchecker}. We then measure how much of the oracle knowledge in each source is covered by these claims.

\noindent\textbf{Results.} 
Figure~\ref{fig:analysis_1} shows that even the best-performing method fails to preserve the majority of oracle knowledge contained in raw experience, suggesting that much of the knowledge that later becomes critical is already lost during pre-construction. 

\setlength{\intextsep}{0pt}

\begin{wrapfigure}{r}{0pt}
\centering
\includegraphics[scale=1.1]{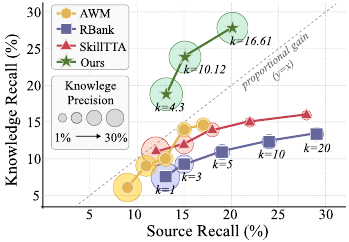}
\caption{Experiments on how effectively existing experience retrieval methods access target-relevant knowledge across different retrieval depths $k$. For Ours, $k$ is the average number of unique sources accessed during navigation.}
\label{fig:analysis_2}
\end{wrapfigure}
\subsection{Preliminary Analysis II: Existing Experience Retrieval Paradigms Provide Limited Access to Target-Relevant Knowledge.}

\paragraph{Setups.}
We next investigate whether existing retrieval paradigms can provide effective access to target-relevant knowledge in past experience. For the analysis, we follow the retrieval setup of each method, where the retrieved sources may consist of pre-constructed skills~\citep{wang2025agent,ouyang2026reasoningbank}, or raw trajectories~\citep{wang2026skillsflytesttimeadaptive}. At the source level, we measure the recall of helpful sources across different retrieval depths \textit{k}. Since retrieving a helpful source does not necessarily indicate how much target-relevant knowledge it actually provides, we further conduct a knowledge-level assessment. Specifically, we compare the unique knowledge contained in the retrieved sources against the full set of unique oracle knowledge for the target task, measuring recall as coverage of required knowledge and precision as the fraction of retrieved knowledge relevant to the target task.

\noindent\textbf{Results.}
Figure~\ref{fig:analysis_2} shows that existing methods achieve low knowledge recall, leaving a substantial portion of target-relevant knowledge inaccessible through retrieval. 
Meanwhile, expanding retrieval with larger \textit{k} does not fully resolve this issue, as marginal gains in knowledge recall are increasingly outweighed by redundant knowledge and irrelevant information, causing precision to drop rapidly. 
This indicates that existing retrieval paradigms provide only limited access to the knowledge needed for the target task, motivating a more targeted form of knowledge access in which agents actively seek what is needed as their understanding of the task's procedural needs evolves during retrieval.

\WFclear

\begin{figure}
    \centering
    \includegraphics[width=1.0\linewidth]{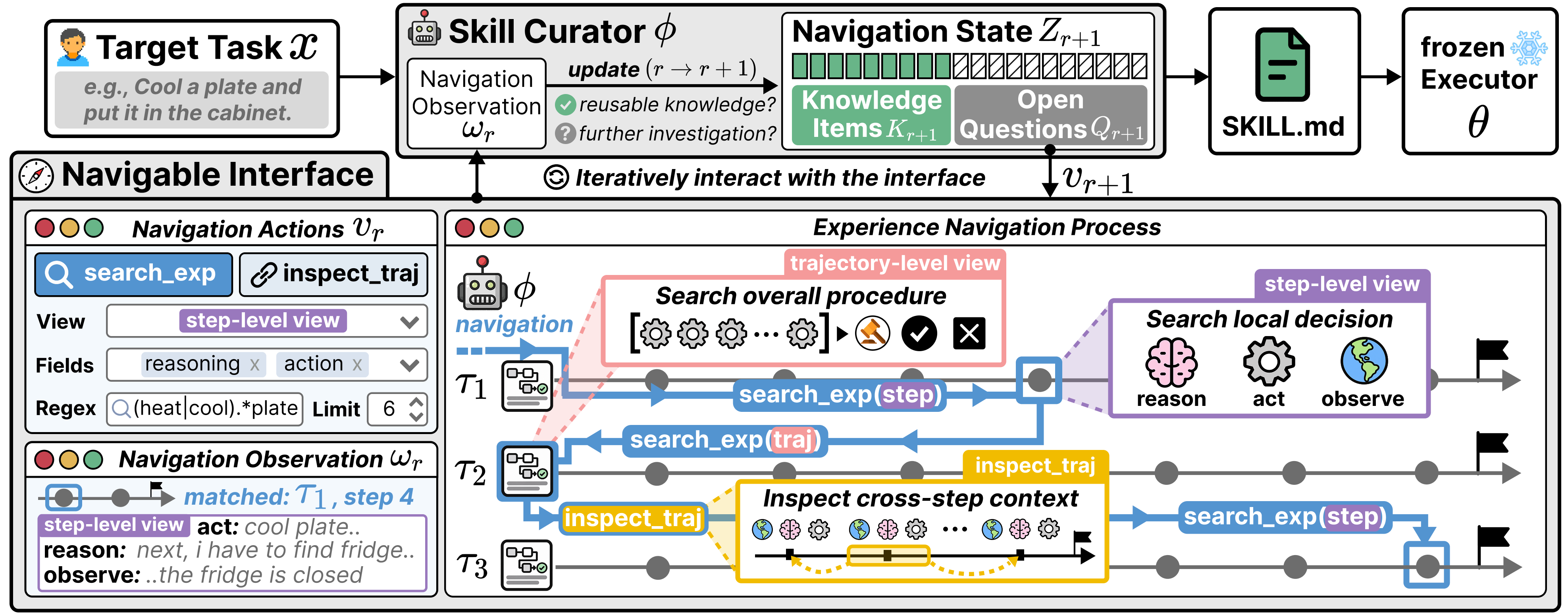}
    \vspace{-5mm}
    \caption{Conceptual overview of \textsc{ExpVoyager}.}
    \label{fig:method}
    \vspace{-5mm}
\end{figure}
\section{ExpVoyager for Dynamic Skill Synthesis}
Motivated by the insights in Section~\ref{section2}, we propose \textbf{\textsc{ExpVoyager}}, a framework for dynamic agent skill synthesis via direct experience navigation.
We present an overview of \textsc{ExpVoyager} in Figure~\ref{fig:method}.

\subsection{Problem Formulation}
Following prior work on learning from experience in LLM agents~\citep{wang2025agent,xia2026skillrlevolvingagentsrecursive,ouyang2026skilloslearningskillcuration}, we study how an agent can reuse its own experience from previous task attempts to solve a new target task.
The available agent experience is represented as a collection of trajectories \(E=\{\tau_i\}_{i=1}^{N}\), where each trajectory \(\tau_i\) contains a task instruction, a sequence of observations and actions, and its execution outcome. 
Given the agent experience \(E\) and a target task instance \(x\), the downstream task is to generate actions that successfully complete \(x\) through interaction with the environment. 
While the broader objective is successful task execution, our focus is on building a curator \(\phi\) to generate a target-conditioned skill \(S_{E,x}\) from \(E\) that provides task-time guidance for completing \(x\) and using it as additional context for a frozen executor \(\theta\):
\[
S_{E,x}=\phi(E,x),
\qquad
a_t \sim \theta\!\left(\cdot \mid x,h_t,S_{E,x}\right).
\tag{1}
\]
where \(h_t=(a_0,o_1,\ldots,a_{t-1},o_t)\) denotes the executor’s interaction history for solving \(x\) at step \(t\).

\subsection{Building Navigable Interface over Agent Experience}

Our goal is to enable more targeted use of past experience by giving the agent active control over how it accesses the relevant knowledge for the current target. 
To achieve this, we equip the skill curator with a \textit{\textbf{Navigable Interface}} over $E$ that exposes past execution through multiple views at different resolutions. 
This is motivated by the intuition that procedural knowledge needed for a target task often resides in different aspects of a trajectory. 
For example, an overall action sequence can reveal the high-level procedure followed to complete a task, whereas understanding a failed operation may require examining the recorded reasoning to identify the condition behind the chosen action. 
Rather than presenting each trajectory as a single retrieval unit, our interface lets the curator inspect evidence at the resolution required by the current investigation and expand to broader context when needed.

\noindent\textbf{Experience Views.} To support such access, we organize each trajectory into complementary knowledge views that preserve the relationship between context, decision, and consequence, without modifying the knowledge contained in the underlying raw experience.

\begin{itemize}[leftmargin=1.5em, labelsep=0.5em, itemsep=3pt, topsep=3pt]
    \item \textbf{Trajectory-level views:}
    Capture the overall execution through the ordered action sequence, execution outcome, and task-specific metadata provided by the environment. These views provide a compact picture of the procedure followed across the episode, helping the curator identify broader strategies, success/failure patterns, or execution stages worth further investigation.

    \item \textbf{Step-level views:}
    Capture individual decisions through the observation, recorded reasoning, executed action, and immediate result, together with surrounding execution context. These views expose the local conditions and immediate consequences of a decision, helping the curator identify conditions associated with different action outcomes.
\end{itemize}

\noindent\textbf{Navigation Actions.} We formulate experience navigation over these views as an iterative decision-making process in which the curator selects both an operation and its arguments at each round. 
Specifically, the curator can invoke \textbf{\texttt{search\_exp}} to locate new experience records across different trajectories by specifying the level of view, fields, a regular-expression pattern, and a result limit as arguments. 
The selected fields and regex pattern determine where matching occurs, while each match is returned as a complete step- or trajectory-level record to preserve the execution context needed for interpretation. 
At the step level, this includes the observation, reasoning, action, and result; at the trajectory level, the action sequence, execution outcome, and task metadata. Each returned record also retains a reference to its source trajectory. When broader procedural context is needed, the curator can invoke \textbf{\texttt{inspect\_traj}} on this reference to return the full source trajectory chronologically. 
At round \(r\), the curator selects an operation and its arguments as a navigation action \(\nu_r\) through the navigable interface \(\mathcal{I}\) over \(E\), receiving the corresponding navigation observation \(\omega_r=\mathcal{I}(E,\nu_r)\).
Based on \(\omega_r\), the curator determines its next operation and arguments, repeatedly moving between localized records and full trajectory context as needed.

\subsection{Guiding Experience Navigation via State Management}
While the navigable interface gives the agent active control over how it accesses past experience, effective navigation requires more than just observing relevant experience.
In particular, the returned experience often capture what happened under source-specific conditions, requiring the curator to interpret what those observations imply for the current target before deciding how to proceed with further navigation.
Moreover, knowledge derived from newly accessed records can refine or revise earlier interpretations and thereby change what remains to be investigated.
To address this challenge, we incorporate a \textbf{\textit{Navigation State} \(Z\)} into the curator’s navigation process, which is updated after each navigation round to reflect what procedural knowledge has been established for the target and what remains to be investigated, so that it can guide the subsequent navigation action:
\[
\left(Z_{r+1},\nu_{r+1}\right)
=
\phi_{\mathrm{nav}}
(Z_r,\omega_r;x),
\qquad
\text{where } Z_r=(K_r,Q_r).
\tag{2}
\] 
Here, \(K_r\) represents the knowledge items curated for the current target, while \(Q_r\) represents the open questions to be investigated through further navigation.

\noindent\textbf{Updating Knowledge Curated from Prior Navigation.}
Rather than simply storing the navigation observations, \(K_r\) accumulates target-relevant knowledge across navigation rounds.
Given the target \(x\), the current state \(Z_r\), and newly accessed navigation observation \(\omega_r\), the curator generates updated knowledge items \(K_{r+1}\) by distinguishing what observations remain specific to the source experience, what procedural relations can transfer to the current target, and what conditions still require verification during execution.
For example, observing that an object was found at a particular location in a past trajectory may support a procedural knowledge for locating and moving the object, but does not establish that the object occupies the same location in the current environment.
The resulting knowledge is organized by its role in execution, while newly accessed \(\omega_r\) can refine the conditions attached to existing knowledge or revise earlier judgments about what transfers to the current target.

\noindent\textbf{Open Questions for Guiding Further Navigation.}
\(Q_{r}\) turns gaps in the current knowledge into concrete objectives for further navigation, focusing on uncertainties whose resolution can make the current knowledge more complete. It is initialized with procedural questions derived from \(x\) and evolves as navigation proceeds: questions supported by newly accessed records can be resolved or revised, while newly identified gaps can be added for subsequent investigation. The curator selects a useful open question and chooses the navigation action and arguments that can help resolve it.

\subsection{Synthesizing Skill for Frozen Executor}

After navigation, the curator uses the resulting \(Z\) to synthesize a \texttt{skill.md} for the frozen executor \(\theta\). Starting from \(Z_0\), initialized from \(x\), the curator iterates the navigation process over \(R\) rounds until further investigation is no longer useful or the navigation budget is exhausted. The curator then generates the final skill as \(S_{E,x}=\phi_{\mathrm{synth}}(Z_R,x)\). The resulting skill turns the accumulated procedural knowledge into tailored execution guidance, retaining unresolved target conditions as execution-time checks and failure lessons as cautions or recovery guidance at relevant stages.
\begin{table*}[t]


\newcommand{\qwenlogo}{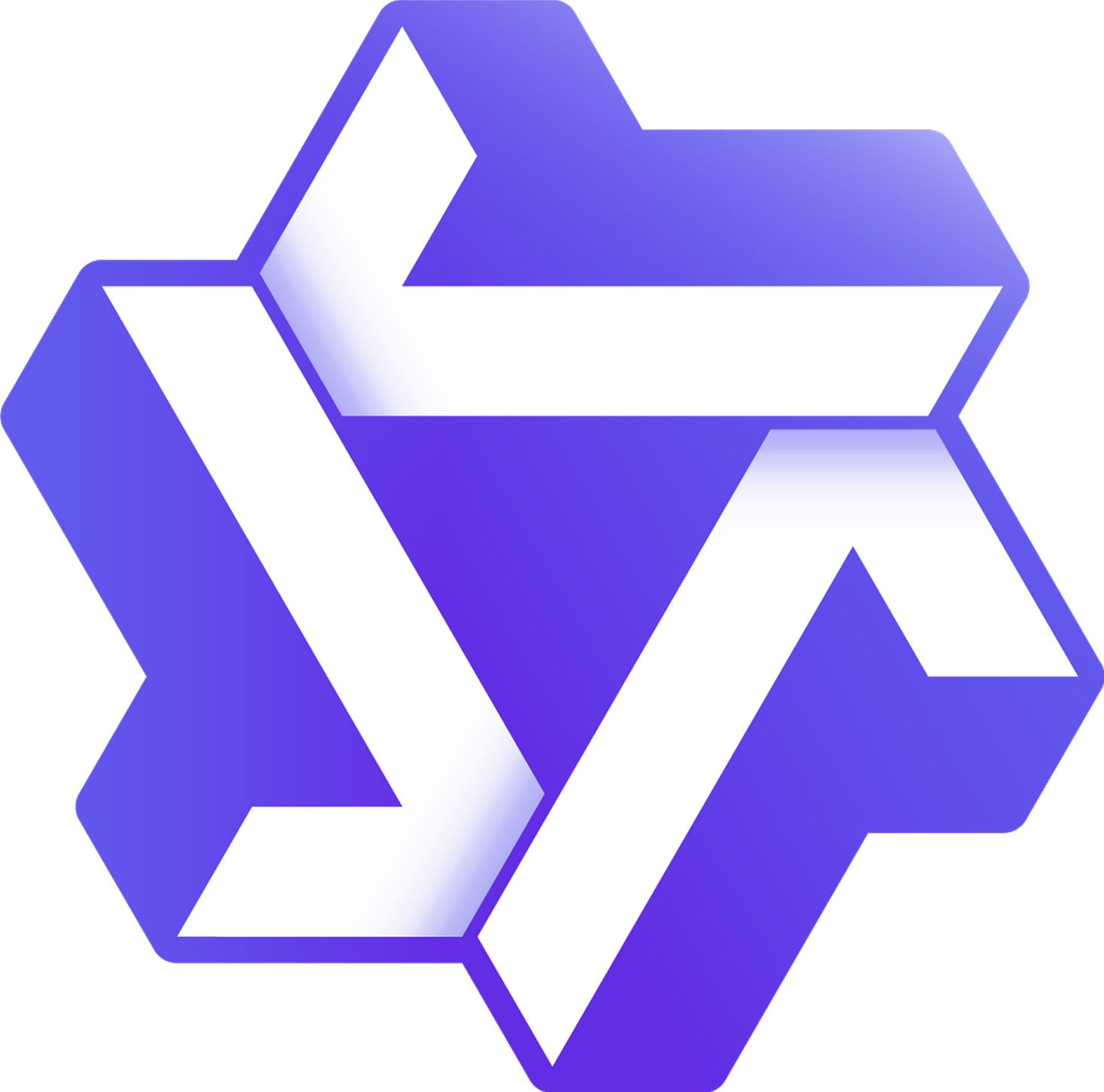}
\newcommand{\googlelogo}{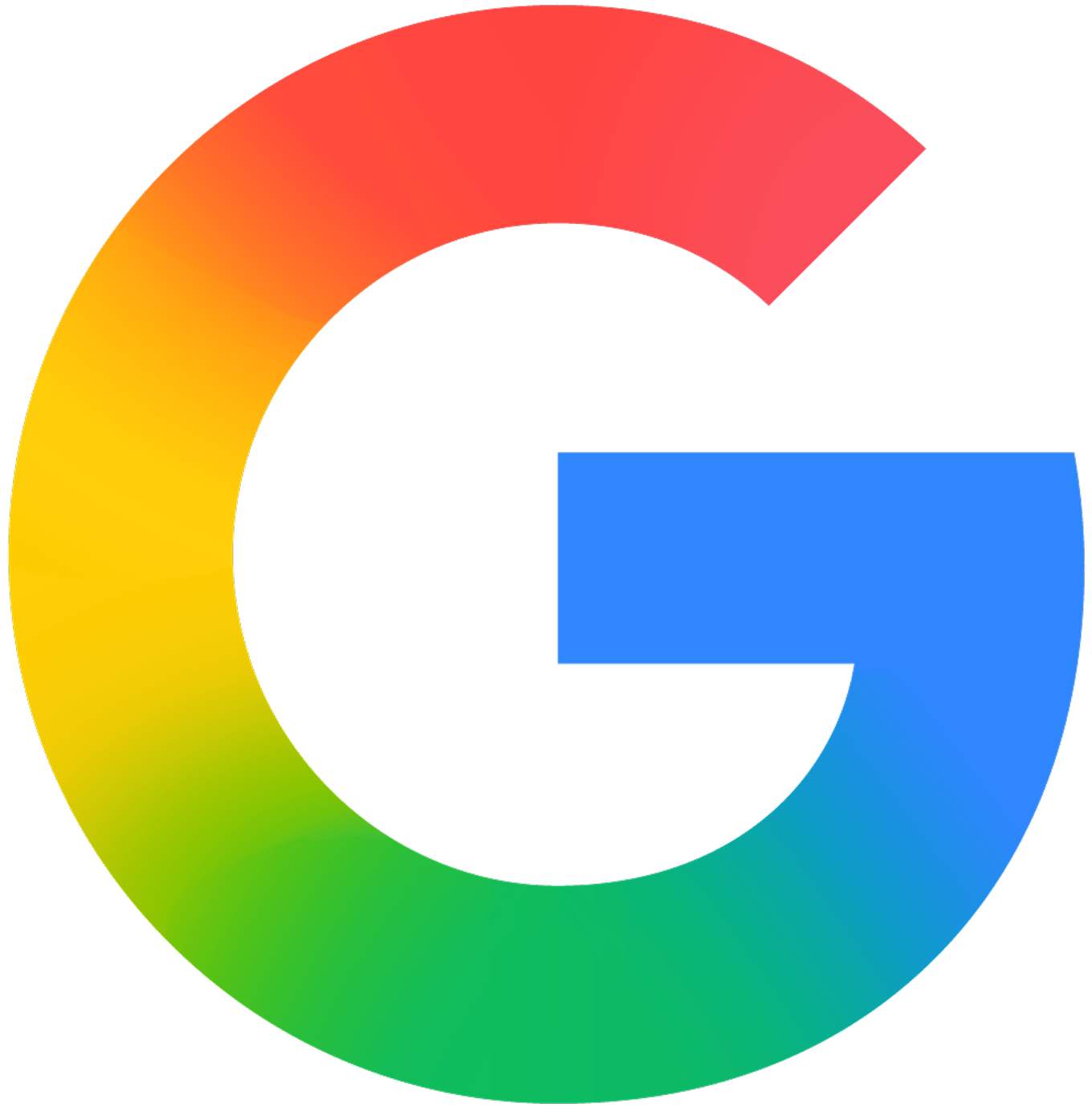}


\newcommand{\benchscore}[2]{%
  \makebox[3.5em][c]{%
    #1\raisebox{-0.35ex}{%
      \scriptsize\textcolor[gray]{0.52}{#2}%
    }%
  }%
}

\newcommand{\benchbestscore}[2]{%
  \makebox[3.5em][c]{%
    \textbf{#1}\raisebox{-0.35ex}{%
      \scriptsize\textcolor[gray]{0.52}{#2}%
    }%
  }%
}

\newcommand{\benchstep}[1]{%
  \makebox[3.5em][c]{#1}%
}

\newcommand{\benchbeststep}[1]{%
  \makebox[3.5em][c]{\textbf{#1}}%
}

\newcommand{\executorheader}[3]{%
  \itshape
  Executor $\theta$:\hspace{0.45em}%
  \raisebox{#3}[0pt][0pt]{%
    \includegraphics[height=1.90ex]{#1}%
  }%
  \hspace{0.35em}%
  #2%
}


\centering
\setlength{\tabcolsep}{2.5pt}
\renewcommand{\arraystretch}{1.1}

\caption{Main results on ALFWorld, WebShop, and ScienceWorld, using Qwen3.5-9B as curator $\phi$.}
\label{tab:main_results}
\vspace{-3mm}

\resizebox{\textwidth}{!}{%
\begin{tabular}{
@{}
l
cc
@{\hspace{3pt}}
ccc
@{\hspace{3pt}}
ccc
@{}
}
\toprule

\multirow{2}{*}{\textbf{Methods}}
&
\multicolumn{2}{c}{\textbf{ALFWorld}}
&
\multicolumn{3}{c}{\textbf{WebShop}}
&
\multicolumn{3}{c}{\textbf{ScienceWorld}}
\\

\cmidrule(lr){2-3}
\cmidrule(lr){4-6}
\cmidrule(lr){7-9}

&
\textbf{SR} ($\uparrow$)
&
\textbf{Steps} ($\downarrow$)
&
\textbf{SR} ($\uparrow$)
&
\textbf{Score} ($\uparrow$)
&
\textbf{Steps} ($\downarrow$)
&
\textbf{SR} ($\uparrow$)
&
\textbf{Score} ($\uparrow$)
&
\textbf{Steps} ($\downarrow$)
\\

\midrule


\rowcolor[gray]{0.92}
\multicolumn{9}{c}{%
  \executorheader{\qwenlogo}{Qwen3.5-9B}{-0.24ex}
}
\\[-0.15ex]

Base (ReAct)~\citep{yao2023react}
& \benchscore{41.4}{1.3}
& \benchstep{34.1}
& \benchscore{19.6}{0.5}
& \benchscore{50.3}{0.4}
& \benchstep{7.1}
& \benchscore{27.0}{1.3}
& \benchscore{37.2}{1.2}
& \benchstep{11.5}
\\

+ AWM~\citep{wang2025agent}
& \benchscore{45.6}{1.8}
& \benchstep{35.1}
& \benchscore{24.2}{0.4}
& \benchscore{52.0}{1.3}
& \benchstep{7.9}
& \benchscore{32.6}{1.7}
& \benchscore{42.7}{1.0}
& \benchstep{8.4}
\\

+ RBank~\citep{ouyang2026reasoningbank}
& \benchscore{43.5}{2.2}
& \benchstep{33.9}
& \benchscore{20.3}{1.4}
& \benchscore{41.6}{1.7}
& \benchstep{9.9}
& \benchscore{29.6}{0.6}
& \benchscore{34.1}{0.6}
& \benchbeststep{7.9}
\\

+ Trace2Skill~\citep{ni2026trace2skilldistilltrajectorylocallessons}
& \benchscore{45.4}{1.7}
& \benchstep{34.3}
& \benchscore{22.6}{0.7}
& \benchscore{45.7}{1.0}
& \benchstep{9.1}
& \benchscore{31.9}{1.3}
& \benchscore{37.6}{0.7}
& \benchstep{9.8}
\\

+ SkillTTA~\citep{wang2026skillsflytesttimeadaptive}
& \benchscore{51.1}{2.8}
& \benchstep{30.4}
& \benchscore{19.2}{1.3}
& \benchscore{34.3}{1.8}
& \benchstep{10.9}
& \benchscore{29.6}{0.6}
& \benchscore{40.7}{0.7}
& \benchstep{11.3}
\\

+ \textbf{\textsc{ExpVoyager}} (Ours)
& \benchbestscore{63.7}{2.2}
& \benchbeststep{25.9}
& \benchbestscore{28.7}{0.6}
& \benchbestscore{61.2}{1.7}
& \benchbeststep{6.9}
& \benchbestscore{37.4}{1.3}
& \benchbestscore{47.0}{0.9}
& \benchstep{11.1}
\\

\midrule


\rowcolor[gray]{0.92}
\multicolumn{9}{c}{%
  \executorheader{\googlelogo}{Gemma4-31B}{-0.20ex}
}
\\[-0.15ex]

Base (ReAct)~\citep{yao2023react}
& \benchscore{53.9}{0.6}
& \benchstep{29.2}
& \benchscore{24.7}{0.7}
& \benchscore{48.4}{0.9}
& \benchstep{7.7}
& \benchscore{48.2}{3.4}
& \benchscore{57.4}{0.5}
& \benchstep{14.9}
\\

+ AWM~\citep{wang2025agent}
& \benchscore{57.0}{0.9}
& \benchstep{27.5}
& \benchscore{30.6}{0.6}
& \benchscore{51.8}{0.4}
& \benchstep{7.9}
& \benchscore{52.2}{1.1}
& \benchscore{59.1}{0.3}
& \benchstep{12.7}
\\

+ RBank~\citep{ouyang2026reasoningbank}
& \benchscore{57.9}{1.5}
& \benchstep{28.8}
& \benchscore{29.8}{0.7}
& \benchscore{51.1}{1.1}
& \benchstep{7.1}
& \benchscore{50.0}{1.1}
& \benchscore{57.7}{1.3}
& \benchstep{12.2}
\\

+ Trace2Skill~\citep{ni2026trace2skilldistilltrajectorylocallessons}
& \benchscore{55.2}{2.4}
& \benchstep{25.4}
& \benchscore{29.4}{0.3}
& \benchscore{50.8}{1.0}
& \benchstep{6.6}
& \benchscore{51.9}{0.6}
& \benchscore{58.4}{0.4}
& \benchstep{11.8}
\\

+ SkillTTA~\citep{wang2026skillsflytesttimeadaptive}
& \benchscore{62.5}{1.6}
& \benchstep{24.7}
& \benchscore{31.3}{1.2}
& \benchscore{57.1}{1.7}
& \benchstep{6.8}
& \benchscore{49.6}{1.7}
& \benchscore{56.4}{0.7}
& \benchstep{13.3}
\\

+ \textbf{\textsc{ExpVoyager}} (Ours)
& \benchbestscore{69.6}{0.1}
& \benchbeststep{22.0}
& \benchbestscore{37.9}{0.8}
& \benchbestscore{66.7}{1.0}
& \benchbeststep{6.1}
& \benchbestscore{55.6}{1.1}
& \benchbestscore{63.7}{1.2}
& \benchbeststep{11.3}
\\

\bottomrule
\end{tabular}%
}

\vspace{-4mm}
\end{table*}
\section{Experiment}

\subsection{Experimental Setup}
We conduct experiments on three popular agent benchmarks across diverse domains, including ALFWorld~\citep{shridhar2021alfworld}, WebShop~\citep{yao2022webshop}, and ScienceWorld~\citep{wang-etal-2022-scienceworld}, covering household interaction, online shopping, and scientific reasoning in interactive environments. 
We evaluate effectiveness (Success Rate) and efficiency (Steps), with specific metrics varying for each dataset. 
For comparison, we consider three categories of baselines: \textbf{(1)} \textbf{base ReAct}~\citep{yao2023react} \textbf{agent without skills}; \textbf{(2) pre-constructed skill} baselines, including AWM~\citep{wang2025agent}, ReasoningBank (RBank)~\citep{ouyang2026reasoningbank}, and Trace2Skill~\citep{ni2026trace2skilldistilltrajectorylocallessons}; and \textbf{(3) test-time skill synthesis} baseline SkillTTA~\citep{wang2026skillsflytesttimeadaptive}. 
For fair evaluation, all baselines and \textsc{ExpVoyager} use the same backbone models within each configuration for both the downstream task execution and skill construction. 
Please refer to Appendix for full descriptions of the datasets~\ref{Datasets}, baselines and evaluation protocols~\ref{Baselines}, implementation details~\ref{implementation}, and prompts~\ref{prompts}.

\subsection{Results: Effectiveness and Versatility of \textsc{ExpVoyager}}
\noindent\textbf{\textsc{ExpVoyager} helps agents reuse past experience through dynamic skills.}
We first evaluate \textsc{ExpVoyager} in an offline setting, where a fixed set of source trajectories is available as past experience, and compare it against the base agent and existing baselines across ALFWorld, WebShop, and ScienceWorld. 
As shown in Table~\ref{tab:main_results} , \textsc{ExpVoyager} consistently improves the performance of the base agent and outperforms existing approaches across all three benchmarks in terms of task success and benchmark-specific scores.
\textsc{ExpVoyager} also reduces the number of execution steps, indicating that the synthesized skills not only improve task completion but also guide the agent toward more efficient execution.
These results suggest that dynamically navigating the experience space and synthesizing actionable skills for the current task provides effective guidance for downstream agent execution, allowing past execution experience to be reused in a task-relevant manner.

\begin{wraptable}{r}{0.585\textwidth}
\vspace{-\intextsep}
\centering


\newcommand{\qwenlogo}{assets/image_1022.png}
\newcommand{\googlelogo}{assets/image_1023.png}
\newcommand{\gptlogo}{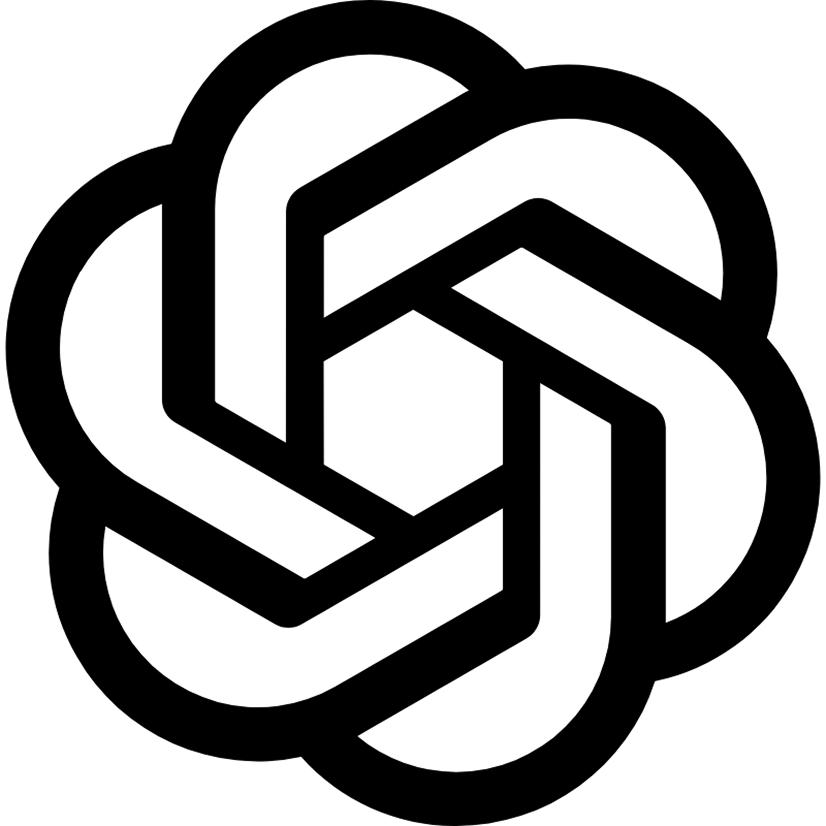}


\newcommand{\score}[1]{%
  #1%
}

\newcommand{\bestscore}[1]{%
  \textbf{#1}%
}

\newcommand{\alfhead}[1]{%
  \hspace{0.8pt}#1\hspace{-0.8pt}%
}

\newcommand{\alfval}[1]{%
  \hspace{2.2pt}#1\hspace{-2.2pt}%
}

\newcommand{\webval}[1]{%
  \hspace{1.3pt}#1\hspace{-1.3pt}%
}

\newcommand{\curatorname}[4]{%
  \raisebox{#3}[0pt][0pt]{%
    \includegraphics[height=#4]{#1}%
  }%
  \hspace{0.25em}%
  #2%
}


\caption{
Performance comparison of diverse curator backbones with
\textsc{ExpVygr} using a Gemma4-31B executor $\theta$.
}
\label{tab:curator_backbone}
\vspace{-2mm}

\scriptsize
\setlength{\tabcolsep}{2.0pt}
\renewcommand{\arraystretch}{0.95}

\resizebox{\linewidth}{!}{%
\begin{tabular}{
@{}
l
l
@{\hspace{1.5pt}}
c@{\hspace{0.5pt}}c
@{\hspace{3pt}}
c@{\hspace{0.5pt}}c
@{\hspace{3pt}}
c@{\hspace{0.5pt}}c
@{}
}
\toprule

\multirow{2}{*}{\textbf{Methods}}
&
\multirow{2}{*}{\textbf{Curator} $\phi$}
&
\multicolumn{2}{c}{\textbf{ALFWorld}}
&
\multicolumn{2}{c}{\textbf{WebShop}}
&
\multicolumn{2}{c}{\textbf{SciWorld}}
\\

\cmidrule(lr){3-4}
\cmidrule(lr){5-6}
\cmidrule(lr){7-8}

&
&
\alfhead{\textbf{SR}}
&
\alfhead{\textbf{Steps}}
&
\textbf{SR}
&
\textbf{Steps}
&
\textbf{SR}
&
\textbf{Steps}
\\

\midrule

Base $\theta$
&
--
&
\alfval{\score{53.9}}
&
\alfval{\score{29.2}}
&
\webval{\score{24.7}}
&
\webval{\score{7.7}}
&
\score{48.2}
&
\score{14.9}
\\

\midrule

+ \textsc{ExpVygr}
&
\curatorname{\qwenlogo}{Qwen3.5-9B}{-0.12ex}{1.80ex}
&
\alfval{\bestscore{69.6}}
&
\alfval{\bestscore{22.0}}
&
\webval{\bestscore{37.9}}
&
\webval{\bestscore{6.1}}
&
\bestscore{55.6}
&
\bestscore{11.3}
\\

+ \textsc{ExpVygr}
&
\curatorname{\googlelogo}{Gemma4-31B}{-0.20ex}{1.80ex}
&
\alfval{\bestscore{71.3}}
&
\alfval{\bestscore{21.8}}
&
\webval{\bestscore{38.4}}
&
\webval{\bestscore{5.7}}
&
\bestscore{60.8}
&
\bestscore{11.2}
\\

+ \textsc{ExpVygr}
&
\curatorname{\gptlogo}{GPT-5.4-mini}{-0.20ex}{1.80ex}
&
\alfval{\bestscore{75.9}}
&
\alfval{\bestscore{19.7}}
&
\webval{\bestscore{38.1}}
&
\webval{\bestscore{5.5}}
&
\bestscore{62.1}
&
\bestscore{13.1}
\\

\bottomrule
\end{tabular}%
}

\vspace{-\intextsep}
\end{wraptable}

\noindent\textbf{\textsc{ExpVoyager} is transferable to diverse executor and curator backbone.}
To validate the transferability of our framework, we evaluate \textsc{ExpVoyager} across diverse curator and executor backbones with different model families and scales.
From Table~\ref{tab:main_results} and \ref{tab:curator_backbone}, \textsc{ExpVoyager} consistently improves task performance across different curator--executor combinations, demonstrating its robustness across diverse model configurations.
In particular, a smaller curator can improve the performance of a much larger executor, highlighting that guidance on what knowledge to reuse from past experience can be just as important as the agent's intrinsic task-solving capability.

\setlength{\intextsep}{0pt}
\begin{wrapfigure}{r}{0pt}
\centering
\includegraphics[scale=1.0]{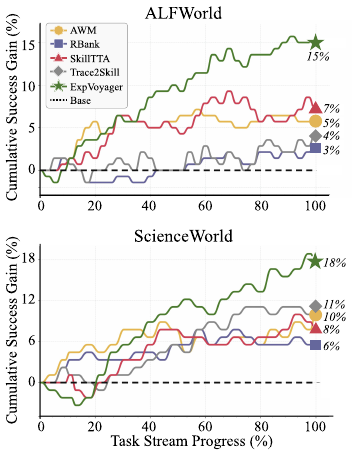}
\vspace{-3mm}
\caption{Online evaluation on ALFWorld and ScienceWorld. The x-axis shows task stream progress, and the y-axis reports cumulative success gain over the base agent.}
\label{fig:online}
\end{wrapfigure}
\noindent\textbf{\textsc{ExpVoyager} drives online self-evolution of agents from accumulating experience.}
While our main experiments assume that a fixed set of past experience is available offline, practical agents may need to continually learn from experience as their experience space is built incrementally during deployment.
To evaluate \textsc{ExpVoyager} in this online setting, we start without a pre-collected experience corpus and sequentially execute the test task stream, and allowing each method to reuse only the experience collected from earlier tasks in the current online stream.
For fair evaluation, we repeat the experiment over three random task orderings and report the average performance.

As shown in Figure~\ref{fig:online}, \textsc{ExpVoyager} provides limited gains and can even degrade performance when the available experience is limited.
However, as the accumulated experience becomes broader, \textsc{ExpVoyager} continues to improve while baseline methods plateau, overcoming its early disadvantage and progressively widening the performance gap.
These results suggest that continual improvement of agent can emerge not only from distilling new experience into a growing store of reusable knowledge for future tasks, but also from accumulating raw experience and navigating it for task-relevant knowledge on demand.

\begin{figure}[H]
\centering

\begin{minipage}[t]{0.495\textwidth}
    \vspace{0pt}
    \centering
    \includegraphics[width=\linewidth]{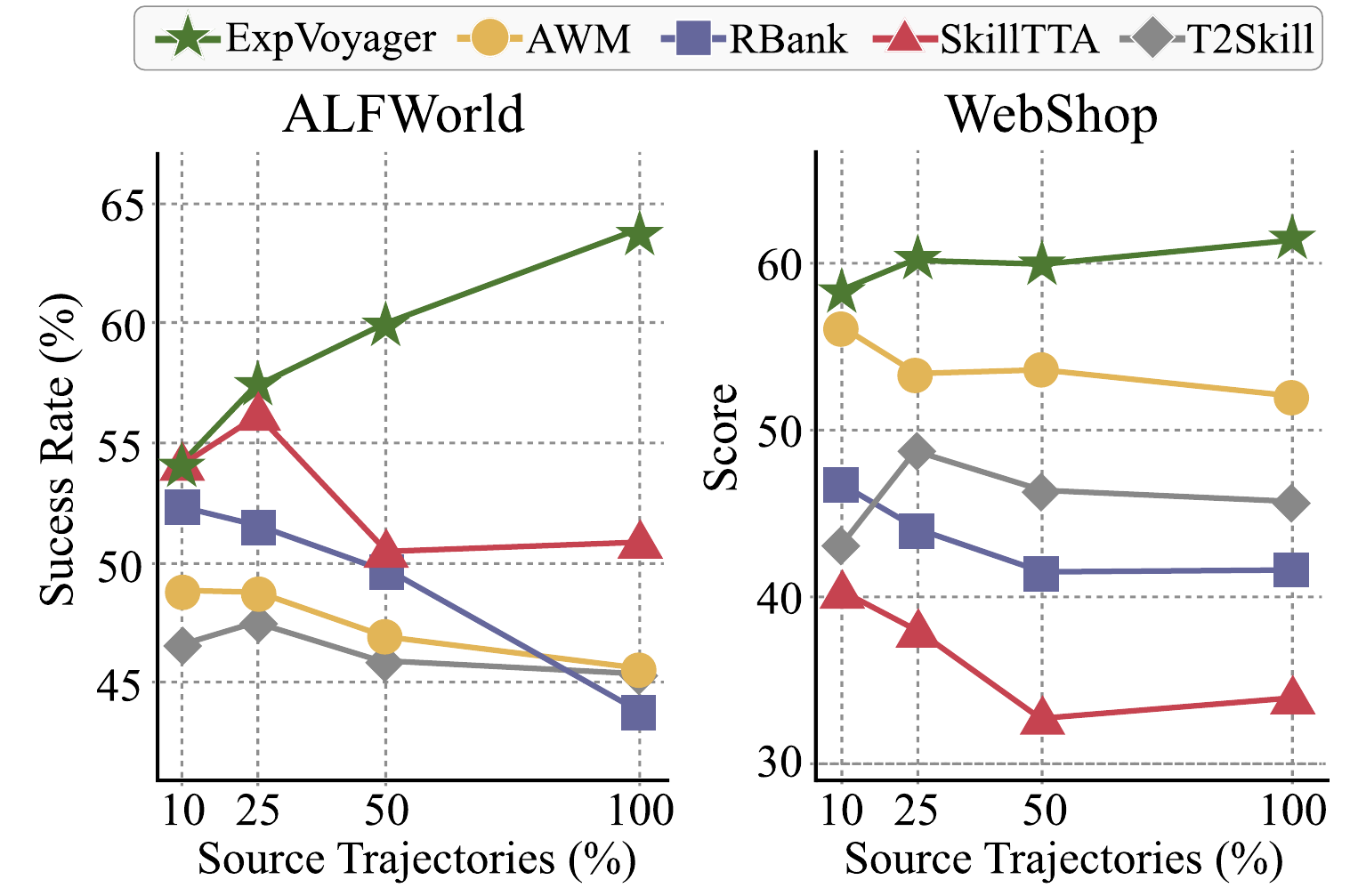}
    \vspace{-6mm}
    \captionof{figure}{Task performance with increasing source experience pool. All methods use the same nested subsets of  1000 souce trajectories (100\%).}
    
    \label{fig:scale}
\end{minipage}
\hfill
\begin{minipage}[t]{0.485\textwidth}
    \vspace{0pt}
    \centering
    \includegraphics[width=\linewidth]{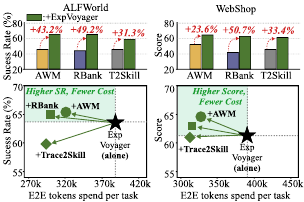}
    \vspace{-5mm}
    \captionof{figure}{Synergy between \textsc{ExpVoyager} and pre-constructed skills. E2E tokens cover the full task-time pipeline; navigation, and execution.}
    \label{fig:synergy}
\end{minipage}

\end{figure}

\noindent\textbf{\textsc{ExpVoyager} scales to larger experience space.}
To further investigate the scaling behavior of \textsc{ExpVoyager} observed in the online setting, we examine how each method scales with increasing amounts of source experience under a controlled offline setup.
Specifically, we construct nested subsets of the source trajectories by progressively increasing the number of available trajectories and provide the same subset to all methods at each scale.
As shown in Figure~\ref{fig:scale}, baseline performance declines as more source experience becomes available, whereas \textsc{ExpVoyager} consistently improves as the source experience grows, with its advantage becoming more pronounced at larger scales.
These results show that simply having more experience available can produce noisier skills for the target task in existing baselines, while \textsc{ExpVoyager} turns a growing experience space into additional gains by effectively identifying task-relevant knowledge from the expanded source experience.




\vspace{2mm}
\noindent\textbf{\textsc{ExpVoyager} can synergize with pre-constructed skills.}
While the previous experiments evaluate \textsc{ExpVoyager} as a standalone approach for dynamic skill synthesis, practical agents may already maintain pre-constructed skills or memories that can provide useful guidance for future tasks.
Such preconstructed knowledge enables efficient reuse across future tasks, but, as discussed in Section~\ref{section2}, abstracting experience into reusable skills in advance risks discarding information that may later become critical for a particular task.
Conversely, synthesizing the on demand skill from raw experience for every task avoids committing to this abstraction in advance, but may repeatedly search for and reconstruct reusable knowledge already captured in pre-constructed skills.

We therefore examine whether \textsc{ExpVoyager} can combine the strengths of both by reusing existing skills as a starting point and dynamically complementing them with on demand knowledge navigation.
Specifically, we compare each \textit{baseline alone}, its \textit{combination with} \textsc{ExpVoyager}, and \textsc{ExpVoyager} \textit{alone}, where the combined setting provides the pre-constructed skill as initial knowledge and lets \textsc{ExpVoyager} navigate the original experience to refine the given skill. 
As shown in Figure~\ref{fig:synergy}, combining \textsc{ExpVoyager} with pre-constructed skills significantly boosts the performance of corresponding baselines. 
When compared with standalone \textsc{ExpVoyager}, the combined variants consistently reduce task-time cost, while also achieving higher task performance with only a single exception.
From the perspective of operating an agent harness system, these results suggest that pre-constructed skills are not required to exhaustively encode all variants of knowledge for future tasks, but can instead serve as reusable foundations that are specialized on demand for each target.

\setlength{\intextsep}{0pt}
\begin{wrapfigure}{r}{0pt}
\centering
\includegraphics[scale=1.2]{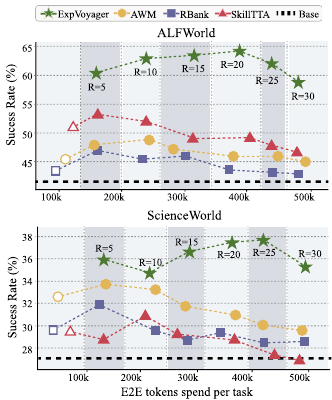}
\vspace{-8mm}
\caption{Comparison with agentic search extensions of baselines under matched experience-access budgets. \(R\) denotes navigation rounds, and E2E token spend cover the full task-time pipeline.}
\label{fig:budget}
\vspace{-3mm}
\end{wrapfigure}
\noindent\textbf{\textsc{ExpVoyager} harnesses experience-access budget more effectively than agentic search.}
 One might wonder whether the gains of \textsc{ExpVoyager} simply come from spending more experience-access budget on repeated access, rather than from its active navigation over the experience space.
To disentangle these effects, we compare \textsc{ExpVoyager} against iterative extensions of existing top-\textit{k} retrieval-based skill approaches.
Specifically, following the iterative retrieval approach  commonly used in agentic search~\citep{jin2025searchr,li-etal-2025-search}, each extension repeatedly performs top-\textit{k} retrieval and refines its experience search query based on the results retrieved in the previous round.
For evaluation under comparable experience-access budgets, we vary the navigation rounds of \textsc{ExpVoyager} and continue each extension of baselines until it reaches a similar token usage to \textsc{ExpVoyager}.

As shown in Figure~\ref{fig:budget}, \textsc{ExpVoyager} consistently outperforms the agentic search extensions under comparable experience-access budgets.
Moreover, increasing the budget yields only marginal gains for baselines and can even hurt performance, whereas \textsc{ExpVoyager} benefits from additional budget up to 20 rounds before it declines with further navigation.
These suggest that effective experience reuse depends not on the amount of access itself, but on targeted access to knowledge that is actually helpful for current task execution.

\vspace{-3.5mm}
\subsection{Analysis: Why and How \textsc{ExpVoyager} works}
\vspace{-3.5mm}
\label{whyandhow}

\vspace{2mm}

\noindent
\begin{minipage}[t]{0.28\textwidth}
    \vspace{0pt}
    \centering

    \captionof{table}{
    Ablation study on ALFWorld. We ablate Navigable Interface and Navigation State to assess their individual contributions.
    }
    \label{tab:component_ablation}

    \vspace{-2mm}

    \setlength{\tabcolsep}{4pt}
    \renewcommand{\arraystretch}{1.20}

    \resizebox{\linewidth}{!}{%
    \begin{tabular}{@{}lc@{}}
    \toprule

    Variants
    & \textbf{SR (\%)}
    \\

    \midrule

    \textbf{\textsc{ExpVygr}} (full)
    & \textbf{63.7}
    \\

    w/o Nav. Interface
    & 55.3
    \\ 

    w/o Nav. State 
    & 58.9
    \\

    \bottomrule
    \end{tabular}%
    }

\end{minipage}%
\hspace{4mm}%
\begin{minipage}[t]{0.71\textwidth}
    \vspace{0pt}
    \centering

    \includegraphics[
        width=\linewidth
    ]{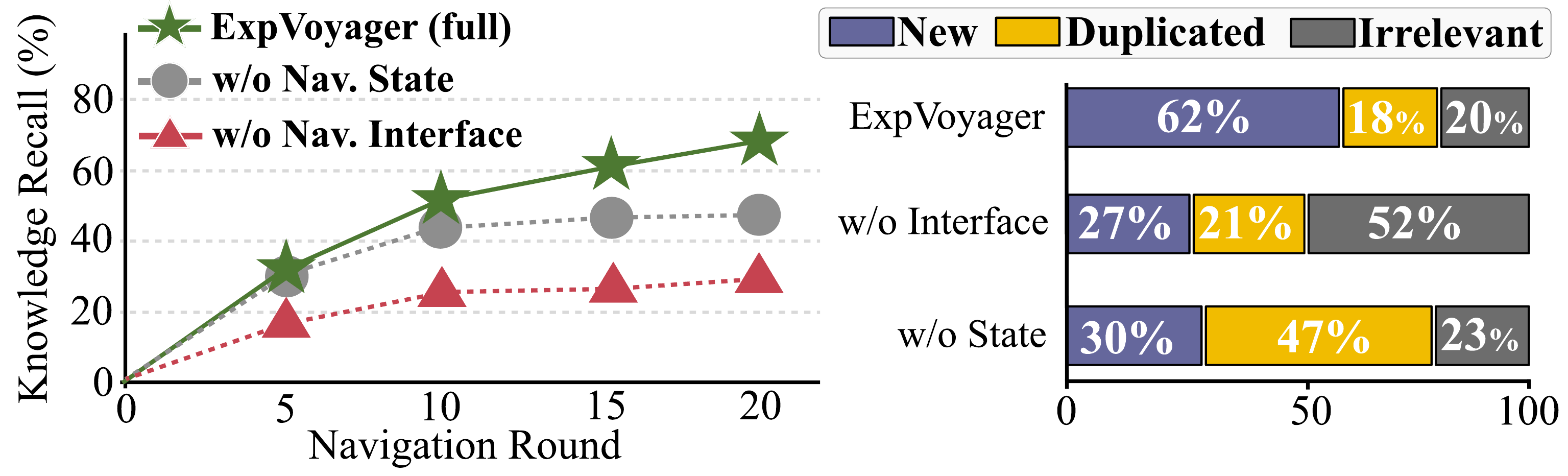}

    \vspace{-2mm}

    \captionof{figure}{
    Effects of two key components on knowledge discovery. (Left) Oracle knowledge recall over navigation rounds. (Right) Composition of knowledge type discovered by each variant.
    }
    \label{fig:ablation_analysis}

\end{minipage}

\vspace{1mm}
\noindent\textbf{Navigable Interface facilitates knowledge access, while Navigation State sustains further discovery.} 
To better understand why \textsc{ExpVoyager} works, we ablate its two main components and analyze the resulting navigation behavior using the oracle knowledge annotations from Section~\ref{section2}. 
Specifically, we compare the knowledge $K_r$ accumulated by \textsc{ExpVoyager} at each navigation round to the oracle knowledge, measuring Knowledge Recall (Figure~\ref{fig:ablation_analysis}, Left) and classifying each knowledge as one of three types: \textit{new knowledge}, \textit{duplicated knowledge}, or \textit{irrelevant knowledge} (Figure~\ref{fig:ablation_analysis}, Right).
In Table~\ref{tab:component_ablation}, removing either component substantially degrades performance.
In Figure~\ref{fig:ablation_analysis}, removing the Navigable Interface lowers Knowledge Recall and increases irrelevant knowledge, whereas removing Navigation State slows recall growth by repeatedly visiting duplicated knowledge.
These results suggest that both components play crucial yet distinct roles in effective experience navigation.


\setlength{\intextsep}{0pt}
\begin{wrapfigure}{r}{0.582\textwidth}
\centering
\includegraphics[width=\linewidth]{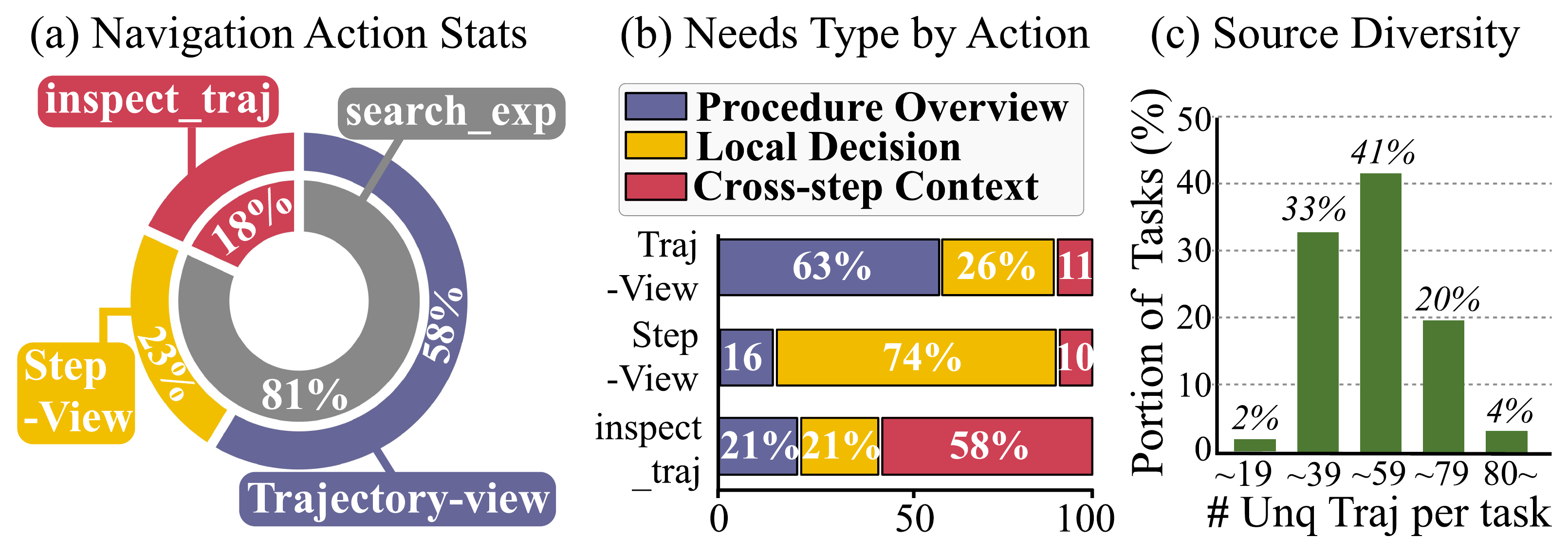}
\vspace{-5mm}
\caption{Experience-access behavior of \textsc{ExpVoyager}. For (b), knowledge types are classified from the open question $Q_r$ at each navigation round.}
\label{fig:resolution}
\end{wrapfigure}
\noindent\textbf{\textsc{ExpVoyager} integrates knowledge across diverse sources by exploring experience at different resolutions.}
To provide an in-depth analysis of how \textsc{ExpVoyager} navigates past experience, we analyze its experience-access behavior during navigation.
Specifically, we report (a) the proportion of selected navigation actions, (b) the types of knowledge sought by each action, and (c) source diversity during navigation.
As shown in Figure~\ref{fig:resolution} (a,b), the curator selects actions according to its information needs, using trajectory-view for procedure overview, step-view for local execution records, and trajectory inspection to understand context across steps.
Figure~\ref{fig:resolution} (c) further shows that \textsc{ExpVoyager} explores diverse procedural knowledge distributed across multiple source trajectories.
These results show that \textsc{ExpVoyager} flexibly accesses experience at different resolutions and across diverse sources as its needs evolve during navigation. We provide detailed setups in Appendix~\ref{anal_why_how}, and case study in Appendix~\ref{case_study}.

\vspace{-3mm}
\section{Related Work}
\vspace{-2mm}

\noindent\textbf{Learning from Experience for Self-Evolving Agents.} 
Recent work studies how agents reuse past experience to improve future decision-making.
Most approaches abstract past trajectories into reusable procedural guidance~\citep{wang2025agent,ouyang2026reasoningbank,ni2026trace2skilldistilltrajectorylocallessons}, while more recent works further learn to curate such guidance from experience~\citep{xia2026skillrlevolvingagentsrecursive,ouyang2026skilloslearningskillcuration}.
Despite these advances, these approaches still construct reusable knowledge before the demands of future tasks are known, which risks discarding knowledge that later becomes critical for a particular task.
While an alternative approach retrieves trajectories and synthesizes a skill for the current task~\citep{wang2026skillsflytesttimeadaptive}, top-$k$ relevance can struggle to capture procedural relevance, as localized cues may also appear in superficially dissimilar trajectories.
In contrast, we reframe skill synthesis as an on demand navigation process that determines what parts of raw experience to inspect and at what resolution, enabling more targeted and fine-grained access to task-relevant knowledge.

\noindent\textbf{Agentic Search and Corpus Interaction.} 
Recently, search paradigms have been shifting from relying on passive retrieval interface toward more active agentic workflows.
Most approaches iteratively refine their search queries based on intermediate reasoning and previously retrieved evidence~\citep{jin2025searchr,li-etal-2025-search,chen2025research}, while more recent work extends agent control to corpus exploration itself, enabling finer inspection of evidence in web corpus~\citep{li2026semanticsimilarityrethinkingretrieval,salemi2026grepseektrainingsearchagents}. 
However, agent experience carries an execution-specific structure where procedural knowledge emerges from relations among context, decision, and consequence. \textsc{ExpVoyager}  exposes theses relations through an interface tailored for experience navigation, allowing agents to move between complementary views both within and across trajectories to access distinct execution semantics.

\vspace{-3mm}
\section{Conclusion}
\vspace{-3mm}

We propose \textsc{ExpVoyager}, a novel framework that reframes agent skill synthesis as dynamic navigation over past experience for targeted and fine-grained access to procedural knowledge on demand. 
Experimental results show that \textsc{ExpVoyager} consistently improves downstream task performance, scales with larger experience spaces, and remains compatible with existing skills under efficient experience access. 
We believe \textsc{ExpVoyager} provides a promising foundation for building self-evolving agents that continuously turn accumulated experience into reusable capabilities.

\section*{AI Use Statement}

We used LLMs to assist with manuscript writing and polishing for clarity and readability, as well as with implementing code for our experiments. The authors reviewed all AI-assisted text and manually verified all AI-generated code before using it in the experiments. The authors take full responsibility for the final content of this work, including all AI-assisted text and code.

\bibliography{iclr2027_conference}
\bibliographystyle{iclr2027_conference}
\clearpage

\newpage
\DoToC
\newpage

\appendix
\section*{Appendix}

\section{Extended Related Work}
\paragraph{Memory Management for Conversational Agents}
Recent work studies how conversational agents preserve information from past user interactions to support consistent and personalized responses across sessions.
Early approaches introduce persistent memory for retaining conversational information~\citep{packer2023memgpt,zhong2024memorybank}, while subsequent work organizes accumulated information into structured or interconnected memories that can be updated as new interactions arrive~\citep{xu2025amem,kang2025memoryosaiagent}.
More recent approaches further learn how memories should be constructed, revised, and utilized from downstream feedback~\citep{tan-etal-2025-prospect, wang2025memalphalearningmemoryconstruction,yan-etal-2026-memory}, or enable query-dependent access to user histories at different levels of abstraction~\citep{seo-lee-2026-p,kim2026bespoke}.
Whereas this line of work primarily uses interaction histories to maintain knowledge about users and prior conversations, \textsc{ExpVoyager} draws on an agent's own task-execution experience to guide new tasks.
Rather than optimizing the memory management policy itself, \textsc{ExpVoyager} navigates past execution records to identify reusable procedural knowledge and compose it into task-specific guidance for the current execution.

\paragraph{Agent Harness Systems.}
Earlier works that augmented foundational language models with retrieval modules~\citep{lewis2020retrieval,pmlr-v162-borgeaud22a, kim-etal-2024-verifiner}, task-specific reasoners~\citep{cheng2023binding,chae-etal-2023-dialogue,seo-etal-2024-unveiling}, specialized tools and APIs~\citep{schick2023toolformer,shen2023hugginggpt,qin2024toolllm}, and program interpreters~\citep{chen2023program,pmlr-v202-gao23f}, can be viewed as precursors to modern agent harnesses.
Building on this broader tradition of augmenting models with external capabilities, recent work has begun to treat the surrounding execution stack itself as a first-class design object, emphasizing the agent harness that equips foundation models with the interfaces, context, and external resources required for effective execution~\citep{openai2026harness,lee2026metaharness}.
One line of work focuses on execution interfaces and environments, designing agent-facing tools and interaction mechanisms that enable models to operate more effectively in complex software and interactive settings~\citep{yang2024sweagent, wang2025openhands}.
Beyond tool access, recent systems increasingly manage the information supplied to the model during execution, maintaining compact context, persistent state, or evolving playbooks to support coherent behavior over long horizons~\citep{zhang2026agentic}.
Reusable agent skills provide another form of harness support by packaging instructions, procedures, and auxiliary resources that can be supplied to the model when relevant to the current task~\citep{anthropic2025agentskills}.
More recent work further makes the harness itself adaptive, optimizing how execution context is constructed or even searching over the implementation of the harness itself using feedback from previous executions~\citep{ye2026meta, lee2026metaharness}.
Whereas these approaches broadly improve the infrastructure and mechanisms surrounding agent execution, \textsc{ExpVoyager} focuses on the procedural knowledge supplied to a given executor for the current task.
Rather than redesigning the execution interface or optimizing the harness itself, \textsc{ExpVoyager} provides a complementary mechanism for adapting the knowledge available to a fixed executor without modifying its underlying harness configuration, allowing the harness to draw on raw experience as an on-demand source of knowledge for agent execution.

\section{Experimental Details}
\paragraph{Default Configuration.}
Unless otherwise specified, we use Qwen3.5-9B as both the skill curator $\phi$ and frozen executor $\theta$, set the navigation budget to $R=20$ rounds, and report results averaged over three random runs.
Except for Tables~\ref{tab:main_results} and~\ref{tab:curator_backbone}, which explicitly vary the curator--executor configuration, all experiments follow this default setting.
For fair comparison, \textsc{ExpVoyager} and all baselines use the same backbone models and  source pool within each configuration.

\subsection{Datasets}
\label{Datasets}

\paragraph{ALFWorld.}
ALFWorld \citep{shridhar2021alfworld} evaluates household task execution through text-based environments constructed by aligning TextWorld with ALFRED. It comprises six task categories---Pick \& Place, Examine in Light, Clean \& Place, Heat \& Place, Cool \& Place, and Pick Two \& Place---across 120 rooms covering kitchens, bedrooms, bathrooms, and living rooms. Given a natural-language goal, an agent navigates the environment and manipulates objects through high-level textual commands, receiving observations after each action. We use 1,000 source trajectories for offline evaluation setting, retaining both successful and unsuccessful attempts as past experience. For evaluation, we follow the official split of ALFWorld and use all 140 instances as our test set, which are separate from the source instances. Each evaluation episode is limited to 50 execution steps.

\paragraph{WebShop.}
WebShop \citep{yao2022webshop} is an interactive shopping benchmark built from approximately 1.18 million Amazon products. Agents use \texttt{search[query]} and \texttt{click[button]} actions to browse products, inspect their details, select options, and complete a purchase that satisfies the user's requirements. Each episode receives a reward between 0 and 1 based on the purchased item's agreement with the requested product type, attributes, options, and price constraint. To construct our source trajectory pool, we sample 1,000 training instructions without replacement using a fixed random seed of 42 and collect one agent rollout per instruction. We retain trajectories with both full and partial or unsuccessful outcomes and follow the official test split, evaluating on all 500 held-out test instructions. Each evaluation episode is limited to 15 execution steps.

\paragraph{ScienceWorld.}
ScienceWorld \citep{wang-etal-2022-scienceworld} assesses scientific reasoning through interactive experiments in a text-based simulated environment. Its 30 task types span 10 science topics, including changes of state, electrical conductivity, chemical mixtures, and plant growth. In the provided environment, agents should translate a goal instruction into a sequence of actions, such as navigating between locations, manipulating materials, and using scientific instruments, while interpreting feedback from the environment. Following the official setting of AgentBoard \citep{ma2024agentboard}, we adopt its fixed evaluation set of 90 ScienceWorld instances. Each evaluation episode is limited to 30 execution steps. We measure performance using the environment's official score, which rewards partial credit for completing required and optional subgoals, normalized to $[0,1]$ and averaged across evaluation instances. We construct our source pool by sampling 500 distinct task--variation pairs from the official training split using a fixed random seed of 42, approximately preserving the task-family proportions of the evaluation set subject to availability. 

\subsection{Baselines}
\label{Baselines}
We compare against four representative baselines that reuse agent experience through workflow retrieval~\citep{wang2025agent}, reasoning memory~\citep{ouyang2026reasoningbank}, global skill consolidation~\citep{ni2026trace2skilldistilltrajectorylocallessons}, and task-specific skill synthesis~\citep{wang2026skillsflytesttimeadaptive}. For fair evaluation, all methods use the same source trajectory pools and are evaluated on the same held-out tasks. Each method constructs its memory or skills from the source experience according to its own procedure.

\paragraph{AWM.}
Agent Workflow Memory (AWM)~\citep{wang2025agent} extracts reusable workflows from successful agent trajectories and provides them as procedural guidance for subsequent tasks. For a fair comparison, we follow the retrieval-based AWM setting used in ReasoningBank~\citep{ouyang2026reasoningbank}, retrieving workflows from a preconstructed memory bank rather than inducing a new workflow for each target task. We use Qwen3-Embedding-8B across all three benchmarks to embed the target's task type and objective, as well as the stored workflow content. Candidate workflows are ranked by cosine similarity across the entire workflow bank for the corresponding benchmark. The top-3 workflows are provided to the executor as additional context.

\paragraph{ReasoningBank.}
ReasoningBank~\citep{ouyang2026reasoningbank} distills reusable reasoning strategies from both successful and failed trajectories, and stores them in a searchable memory bank. We follow the official implementation for memory extraction and retrieval, using the source trajectories to construct the memory bank. For all benchmarks, we use Qwen3-Embedding-8B for embedding-based similarity search. Given a target task, the method retrieves the most relevant experience entry and injects its associated reasoning memories into the executor's context. We use top-1 retrieval, which achieved the best performance in the retrieval-size ablation reported in the original paper. 

\paragraph{Trace2Skill.}
Trace2Skill~\citep{ni2026trace2skilldistilltrajectorylocallessons} consolidates lessons from agent trajectories into a reusable skill through a hierarchical map--reduce procedure. Starting from a frozen seed skill, we independently process each source trajectory in the MAP stage to propose structured edits, extracting reusable procedures from successes and corrective checks from failures. These proposals are then merged hierarchically with a fan-in of five, resolving conflicts and removing redundant guidance at each level. The final merged edits are applied deterministically to the seed, producing a single \texttt{SKILL.md} for each benchmark. This skill is shared across all evaluation tasks within the benchmark. To use the same source experience as the other baselines, we reuse the existing source trajectories rather than collecting new rollouts conditioned on the seed skill.

\paragraph{SkillTTA.}
SkillTTA~\citep{wang2026skillsflytesttimeadaptive} synthesizes a temporary skill for each target task from retrieved source trajectories.  
Specifically, we encode source metadata and the visible target context with Qwen3-Embedding-8B across all three benchmarks and retrieve the top-3 source trajectories by cosine similarity for each target, considering both successful and unsuccessful attempts. 
The retrieved trajectories and target context are used to synthesize a task-specific \texttt{SKILL.md} containing applicability conditions, possible failure modes, procedures, and a verification checklist.

\subsection{Implementation Details}

\subsubsection{Details of \textsc{ExpVoyager}}
\label{implementation}
\paragraph{Models and local inference.}
We serve Qwen3.5-9B and Gemma4-31B locally using vLLM with an OpenAI-compatible Chat Completions interface on NVIDIA RTX A6000 GPUs with 48 GB memory. In the six-GPU configuration, Qwen3.5-9B uses six independent single-GPU replicas with tensor parallelism of one, while Gemma4-31B uses three replicas with tensor parallelism of two. Both local models run in BF16 with a context window of 32,768 tokens, a temperature of 1.0, and thinking enabled. We use model-specific reasoning parsers and the Hermes and Gemma4 tool-call parsers for Qwen and Gemma, respectively. For experiments using GPT-5.4-mini as the curator (Table~\ref{tab:curator_backbone}), we access the model through the OpenAI Responses API with reasoning effort set to medium, without explicitly specifying sampling parameters. The corresponding Gemma4-31B executor remains locally served, using two replicas with tensor parallelism of two.

\paragraph{Experience representation.}
We retain source trajectories as chronological execution records and materialize trajectory-level and step-level views in separate JSONL indexes. Trajectory-level records contain the ordered action sequence, episode-level execution outcome. Step-level records contain the observations (before and after action), executor reasoning, executed action, and immediate result. Both views retain a reference to the full source trajectory. 

\paragraph{Navigable Interface.}
The curator invokes \texttt{search\_exp} by specifying the view level, fields, regular-expression pattern, and result limit. We implement case-insensitive regular-expression matching in Python over the selected fields. Matching records are returned in stored order up to the requested limit. Each match returns the complete stored view record and its source reference, even when matching uses only a subset of the fields. Trajectory-level searchable fields comprise the action sequence, and outcome; step-level searchable fields comprise the observation, reasoning, action, and result. The curator invokes \texttt{inspect\_traj} on a source reference to access the chronological trajectory. The underlying reader supports line offsets and limits, with continuation information for longer records.

\paragraph{Native tool calling and navigation control.}
Navigation actions are implemented through native function calling. Tool descriptions and JSON-schema argument specifications are supplied through the request's \texttt{tools} field. The runtime parses the returned \texttt{tool\_calls} and dispatches the corresponding Python function. We set \texttt{parallel\_tool\_calls=False} to execute one navigation operation per action-selection request. Navigation requests use automatic tool selection, whereas initialization and state-update requests explicitly select their respective state-management functions through \texttt{tool\_choice}. The curator may return \texttt{FINISH} after incorporating valid navigation observations into its state. Navigation otherwise continues until the default budget of 20 rounds is exhausted. 

\begin{figure}
    \centering
    \includegraphics[width=1.0\linewidth]{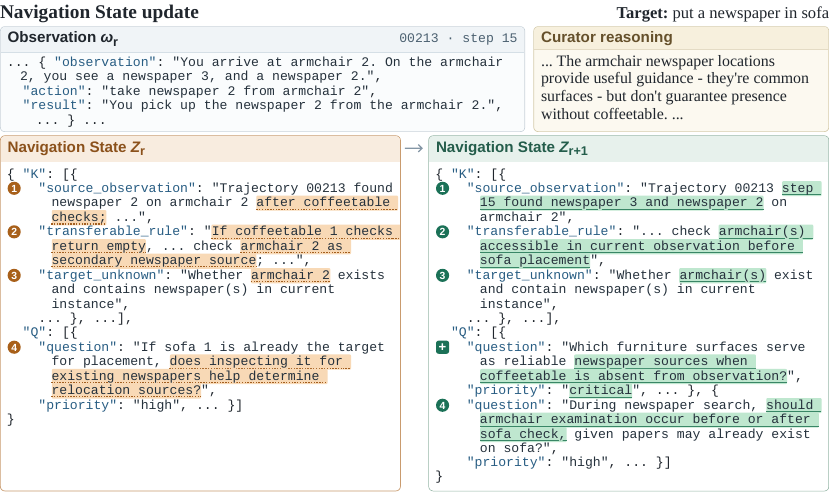}
    \vspace{-5mm}
    \caption{\textbf{Example of a Navigation State update in ALFWorld.} The curator uses navigation observation $\omega_r$ to revise knowledge \(K_r\) and open questions \(Q_r\), while newspaper availability in the target remains unverified. Orange and green highlight prior and updated content, respectively; matching numbers link corresponding content, and \(+\) marks a new question.}
    \label{fig:navigation_state}
    \vspace{-5mm}
\end{figure}
\begin{figure}
    \centering
    \includegraphics[width=1.0\linewidth]{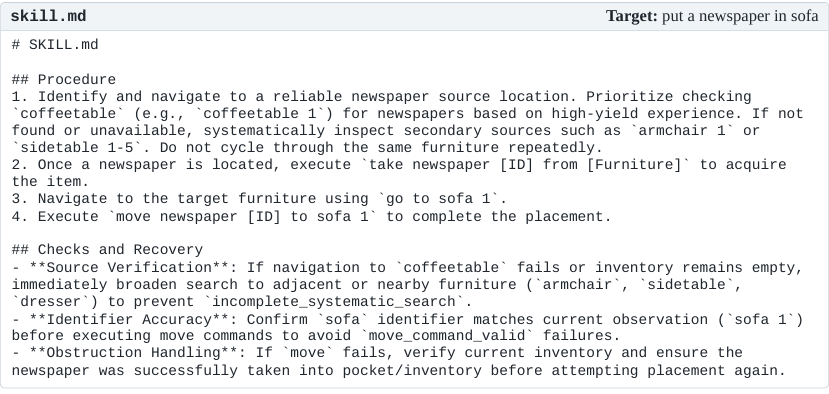}
    \vspace{-5mm}
    \caption{\textbf{Generated \texttt{skill.md} for the same ALFWorld task.} Synthesized after the final navigation round, the skill provides a procedure for locating and placing a newspaper, with checks for source availability, identifier accuracy, and placement recovery. The full generated Markdown is shown.}
    \label{fig:skill}
    \vspace{-5mm}
\end{figure}
\paragraph{Navigation State.}
We represent $Z_r=(K_r,Q_r)$ as a structured JSON object initialized from the target context. Each knowledge item records a source-specific observation, a transferable procedural relation, and conditions that remain to be verified during execution. Items are organized by their role in execution, while open questions retain unresolved procedural uncertainties and their priorities. After receiving a navigation observation, a separate curator request assesses its relevance and updates the knowledge items and open questions. The runtime validates the structured update and maintains references to supporting observations. Subsequent navigation requests receive the target context, current Navigation State, active experience records, and navigation feedback. After the final navigation round, the curator synthesizes \texttt{skill.md} from the resulting state and target context. Figures~\ref{fig:navigation_state} and~\ref{fig:skill} illustrate an example of Navigation State update and a generated \texttt{skill.md}, respectively.
\subsubsection{Prompts}
\label{prompts}
We present the curator prompts for Navigation Action Selection, Navigation State Update, and Final Skill Synthesis. The system and user prompts for Navigation Action Selection (Tables~\ref{tab:navigation_action_selection_prompt} and~\ref{tab:navigation_action_selection_user}) guide the curator to select an open question, choose the appropriate experience view and search fields, and determine when to finish navigation. The native function schemas for \texttt{search\_exp} and \texttt{inspect\_traj} (Tables~\ref{tab:search_exp_tool_schema} and~\ref{tab:inspect_traj_tool_schema}) specify the arguments for searching experience records and inspecting a source trajectory, respectively. 
The system and user prompts for Navigation State Update (Tables~\ref{tab:navigation_state_update_system} and~\ref{tab:navigation_state_update_user}) guide the curator to determine whether the navigation observation answers the current question, distinguish transferable knowledge from facts requiring verification during execution, and update the knowledge items and open questions. The Final Skill Synthesis prompt (Table~\ref{tab:final_skill_synthesis_prompt}) instructs the curator to generate \texttt{skill.md} from Navigation State of the final round, preserving supported procedures, conditional guidance, and relevant checks and recovery strategies.

\subsection{Experimental Details for Preliminary Analyses}
\label{Preliminary}

\paragraph{Oracle skill construction.}
We define \textit{oracle skill} as a target-specific skill whose helpfulness for the corresponding task has been verified through repeated execution, rather than as a unique or exhaustive description of all knowledge that could solve the task.
We use GPT5.6 Sol for candidate skill generation, knowledge annotation, and claim-level evaluation, and Qwen3.5-9B for trajectory collection and execution-based verification. For each target task, we initially collect ten independent execution rollouts and collect additional rollouts as needed to obtain at least four successful and four failed trajectories, ensuring sufficient evidence for both effective procedures and failure-specific checks or recovery strategies. Given the target instruction, initial observation, and collected trajectories with their outcomes, the model synthesizes a candidate skill containing actionable procedures and relevant checks or recovery strategies.
We accept a candidate as an oracle skill only after verifying its helpfulness for the target task through execution. Specifically, we provide each candidate skill to the same frozen executor and conduct three independent executions of the corresponding task. To reduce the influence of execution randomness, we retain a candidate only if all three executions succeed. Otherwise, failed execution trajectories are added to the evidence pool before generating the next candidate. We allow up to 20 candidate versions per target and obtain 80 oracle skills through this iterative generate-then-verify process. Although repeated execution-based verification reduces the influence of execution randomness, successful outcomes can still arise from task-specific shortcuts or other spurious factors. We therefore conduct a manual review of each retained oracle skill and its execution trajectories to confirm that the target requirements are genuinely satisfied and that the skill provides useful procedural guidance without effectively substituting for execution through an overly instance-specific solution. 

\paragraph{Source sampling and oracle knowledge annotation.}
We sample 300 source trajectories from the ALFWorld training split, including both successful and unsuccessful attempts. For each of the 80 oracle skills, we examine every source trajectory. The annotation model receives the target information, the oracle skill, and the source trajectory, including its recorded reasoning, actions, observations, and outcome information. The oracle skill identifies the knowledge needed for the target, while the source trajectory provides the experience supporting that knowledge. An item is annotated only when supporting evidence can be identified in the source itself.
A source is labeled helpful when it contains non-trivial knowledge that can contribute to constructing the corresponding oracle skill. Such knowledge includes actionable procedures, state transitions, preconditions, ordering constraints, failure causes, recovery strategies, and completion checks. Generic interface knowledge, task restatements, superficial object-name overlap, and unsupported speculation are excluded. Failed trajectories can therefore be helpful when they reveal valid partial progress or a concrete failure lesson, while successful trajectories can also   receive a \textit{none} label when they contribute no relevant knowledge. Each annotated item is linked to a supporting excerpt from the source, and sources without qualifying items are labeled \textit{none}. After annotation, manually review all knowledge annotations, including retained knowledge items and \textit{none} labels, to verify that annotated items are supported by the cited source evidence and that the above inclusion and exclusion criteria are applied consistently.

\paragraph{Claim extraction and semantic matching.}
Following the claim-level evaluation protocol of \citet{ru2024ragchecker,seo-etal-2025-mt}, we separate evaluation into claim extraction and semantic matching. We adapt this principle to procedural knowledge in skills and agent experience, assessing the degree of semantic correspondence on a five-point Likert scale.
First, we decompose each evaluated artifact into atomic knowledge claims. An atomic claim expresses one independently assessable procedure, condition, state transition, or verification rule. Statements containing several separable instructions are split, while conditions essential to an instruction's meaning are preserved. Oracle knowledge is hidden during this extraction stage so that claims are extracted from the artifact independently of the desired matches. Each extracted claim is linked to supporting text in the artifact.
Next, we rate the semantic correspondence between each extracted claim and the target's oracle knowledge on a 1--5 Likert scale. The judgment considers how closely the actionable knowledge and its required conditions, ordering constraints, causal relations, and verification requirements align, rather than relying on surface-level textual similarity. Differences in wording or object names are permitted when the claims express the same transferable knowledge, whereas shared keywords or task categories alone are insufficient. A score $s$ is normalized to $[0,1]$ using $(s-1)/4$.

For aggregation, each item contributes only its highest matching score against the comparison set. When measuring oracle knowledge coverage, we use the score of the best-matching claim for each oracle item. When measuring the precision of retrieved knowledge, we use the score of the best-matching oracle item for each retrieved claim. Scores from multiple matches are not summed, so each item contributes at most once to the corresponding metric.

\paragraph{Analysis I: Knowledge preservation in pre-constructed skills.}
We evaluate the pre-constructed skills or memory artifacts of AWM, ReasoningBank, and Trace2Skill. For each target, we compare their extracted atomic claims with the oracle knowledge annotated in the corresponding source trajectories.
We record which source trajectories are used to construct each artifact and use the oracle knowledge annotated in those sources as its reference. For methods that consolidate multiple trajectories, such as AWM~\citep{wang2025agent} and Trace2Skill~\citep{ni2026trace2skilldistilltrajectorylocallessons}, we compare the consolidated artifact against the oracle knowledge from its contributing sources without duplicating the artifact for each source. All methods are evaluated against oracle knowledge annotated over the same source pool.
Knowledge preservation is measured by taking the highest normalized matching score between each oracle item and the artifact's claims, then averaging these scores. This captures how faithfully the source knowledge is retained in the pre-constructed artifacts. We compute preservation scores separately for each target and then macro-average across targets.

\paragraph{Analysis II: Source-level evaluation.}
We evaluate AWM, ReasoningBank, and SkillTTA at retrieval depths $k \in \{1,3,5,10,20\}$, following each method's retrieval setup. AWM retrieves pre-constructed workflows, ReasoningBank retrieves memory entries, and SkillTTA retrieves raw source trajectories. We use Qwen3-Embedding-8B for embedding-based retrieval. Candidate pools are restricted to the sampled source trajectories or artifacts constructed from them, and oracle annotations are used only for evaluation.

For target $t$, let $H_t$ denote the set of source trajectories labeled helpful for that target in the preceding helpfulness annotation. $R_{m,t}^{k}$ denotes the original source trajectories associated with the top-$k$ artifacts retrieved by method $m$. We identify these sources from the records of which trajectories were used to construct each artifact and count repeated source IDs only once. Source recall is defined as
\[
\operatorname{SourceRecall}@k
=
\frac{\text{Number of retrieved helpful sources}}
{\text{Total number of sources annotated as helpful}}.
\]

Here, $k$ counts each method's native retrieval units. For AWM, which consolidates multiple source trajectories into each workflows, one retrieved workflow may correspond to several source trajectories.

\paragraph{Analysis II: Knowledge-level evaluation.}
For each target $t$, we construct a unique knowledge set $K_t$ by removing semantically similar or duplicate items expressing the same knowledge from the oracle annotations over the 300 source trajectories. Items that differ only in wording are represented once, while distinct conditions, ordering requirements, or recovery rules remain separate.
We then extract atomic claims from the actual retrieved content: workflow text for AWM, memory text for ReasoningBank, and raw trajectory content for SkillTTA. Semantically duplicate claims are removed from the retrieved collection to obtain a unique claim set. We apply the same 1--5 scoring and normalization procedure described above.
Let $r_i$ denote the highest normalized matching score between oracle item $i$ and the retrieved claims, and let $p_j$ denote the highest normalized matching score between retrieved claim $j$ and the oracle knowledge. With $N$ oracle items and $L$ unique retrieved claims, we calculate
\[
\operatorname{KnowledgeRecall}@k
=
\frac{1}{N}\sum_{i=1}^{N} r_i,
\qquad
\operatorname{KnowledgePrecision}@k
=
\frac{1}{L}\sum_{j=1}^{L} p_j.
\]

Both $r_i$ and $p_j$ lie in $[0,1]$. Recall measures how faithfully the target's oracle knowledge is covered by the retrieved material, while precision measures how closely the retrieved knowledge aligns with the target's oracle knowledge.
Retrieving a helpful source does not automatically credit all knowledge associated with that source. Matching scores must be supported by the content actually retrieved. Both metrics are computed per target and macro-averaged across targets at each retrieval depth.
\subsection{Experimental Details for Comparison to Agentic Search}
\label{Agentic_search}
To examine whether \textsc{ExpVoyager}'s gains come simply from additional experience access, we compare it against agentic search extensions of retrieval-based baselines. Starting from the target task, each extension repeatedly retrieves top-\(k\) skills or trajectories and refines its search query based on the retrieved results. Processing these results and reasoning about subsequent retrieval queries incur additional tokens, allowing us to compare agentic search through a passive retrieval interface with active navigation under comparable experience-access budgets. Specifically, we operationalize the experience-access budget as the token usage of the entire task-time process before downstream execution, including experience retrieval or navigation, reasoning over the accessed experience, and any method-specific step used to construct the final guidance. For \textsc{ExpVoyager}, this includes navigation-action selection, processing of accessed experience, Navigation State updates, and final skill synthesis. For the baselines, we follow their native task-time pipelines: SkillTTA includes target-specific skill synthesis after retrieval, whereas AWM and ReasoningBank directly reuse their pre-constructed workflows or memories and therefore incur no additional synthesis stage. We vary \textsc{ExpVoyager}'s navigation rounds and continue iterative retrieval for each baseline until this pre-execution token usage approximately matches that of \textsc{ExpVoyager}. Downstream task execution is excluded from the matched budget and is included only when reporting the full end-to-end token cost. From the results, \textsc{ExpVoyager} consistently outperforms these extensions under comparable budgets, suggesting that active navigation makes more effective use of the available task-time computation. However, its performance also declines beyond approximately 20--25 rounds, possibly because continued navigation after sufficient useful experience has been collected introduces redundant or less relevant experience that can interfere with task execution. We present the full results in Table~\ref{tab:extended_budget_alfworld}.

\subsection{Experimental Details for Analysis on why \textsc{ExpVoyager} works}
To better understand the distinct roles of the Navigable Interface and Navigation State, we ablate each component while preserving the remaining \textsc{ExpVoyager} pipeline and navigation setup (Section~\ref{whyandhow}). 

\paragraph{w/o Navigable Interface.} We retain the Navigation State and the same iterative navigation procedure, but remove the experience-specific views and access operations provided by the Navigable Interface. Specifically, similar to \citet{li2026semanticsimilarityrethinkingretrieval}, we expose the raw source trajectories as chronological files and allow the curator to interact with them through general-purpose terminal operations (e.g., grep/rg) over the serialized trajectory content. Unlike the full \textsc{ExpVoyager}, this interface does not explicitly expose the execution-specific structure of agent experience through complementary views. Instead, distinct execution semantics, such as episode-level procedural patterns and local context–decision–consequence relations, remain embedded in the raw trajectory content rather than being surfaced through dedicated views and view-specific fields for targeted access. The matched portions of source trajectories, with their surrounding context determined by the curator’s search and read operations, are then passed to the same Navigation State updater as in the full \textsc{ExpVoyager}, which guides subsequent navigation.

\paragraph{w/o Navigation State.} We retain the full Navigable Interface and the same iterative navigation procedure, but remove the explicit Navigation State. Specifically, instead of maintaining \(Z_r=(K_r,Q_r)\) across rounds, the curator conditions each navigation decision on the target context and the accumulated navigation history, including previous navigation actions and the experience records accessed through the interface. To keep the interaction context bounded across rounds, earlier navigation history is compressed into a simple running summary that preserves the information needed for subsequent decisions. Thus, evidence from earlier rounds remains available in the interaction context, but is not explicitly interpreted and consolidated into target-relevant procedural knowledge and open questions that are revised after each navigation round. This variant therefore preserves the same experience-access operations and navigation process while removing the explicit state management that connects the interpretation of previously accessed experience with what to investigate next.

\subsection{Experimental Details for Analysis on How \textsc{ExpVoyager} works}
\label{anal_why_how}
We analyze ExpVoyager's experience-access behavior on ALFWorld, Webshop, and ScienceWorld with three runs per task. Both the curator and executor use Qwen3.5-9B, and navigation is limited to 20 rounds. The remaining settings follow the main experiments. We collect the navigation actions, open questions, curator reasoning, and accessed source trajectories to examine action selection, information needs, and source diversity.

For Figure~9(a), we group navigation actions into trajectory-view search, step-view search, and trajectory inspection. We aggregate the number of calls across all tasks and runs and compute the proportion of each action. The inner ring shows the proportions of \texttt{search\_exp} and \texttt{inspect\_traj}, while the outer ring further divides \texttt{search\_exp} into trajectory-view and step-view searches.

For Figure~9(b), we classify the information sought at each navigation round using the open question $Q_n$ together with the curator's reasoning for that navigation decision. GPT-5.6 Sol assigns one primary category to each round. \emph{Procedure Overview} covers questions about the overall procedure or ordering of subgoals. \emph{Local Decision} covers questions about a specific action, its execution conditions, or command details. \emph{Cross-step Context} covers questions about dependencies between actions or state changes across multiple steps. When a question involves multiple needs, the primary category is determined by the main information sought, as expressed in the question and accompanying reasoning. We then group these labels by the selected action and report the proportion of each category within each action.

For Figure~9(c), we measure source diversity by counting the unique source trajectories accessed during navigation for each task and run. This includes trajectories accessed through trajectory-view search, step-view search, and trajectory inspection. Multiple records or repeated accesses originating from the same trajectory count as a single source. We group the resulting counts into intervals of 0--19, 20--39, 40--59, 60--79, and 80 or more trajectories, and report the percentage of tasks in each interval averaged over the three runs.
\clearpage

\begin{table*}[t]

\newcommand{\score}[2]{%
  #1\raisebox{-0.35ex}{\scriptsize\color[gray]{0.52}#2}%
}

\newcommand{\bestscore}[2]{%
  \textbf{#1}\raisebox{-0.35ex}{\scriptsize\color[gray]{0.52}#2}%
}

\newcommand{\stepval}[1]{%
  #1%
}

\newcommand{\beststep}[1]{%
  \textbf{#1}%
}

\centering
\setlength{\tabcolsep}{3.5pt}
\renewcommand{\arraystretch}{1.08}

\caption{Results on ALFWorld across different subtask types, using Qwen3.5-9B as the curator $\phi$.}
\label{tab:alfworld}

\resizebox{\linewidth}{!}{%
\begin{tabular}{@{}l*{7}{c}c@{}}
\toprule

\textbf{Methods}
& \textbf{Pick}
& \textbf{Look}
& \textbf{Clean}
& \textbf{Heat}
& \textbf{Cool}
& \textbf{Pick2}
& \textbf{Avg. SR}
& \textbf{Steps}
\\
\midrule


\rowcolor[gray]{0.92}
\multicolumn{9}{c}{%
  \rule{0pt}{2.25ex}\itshape
  Executor $\theta$: Qwen3.5-9B
}
\\

Base (ReAct)~\citep{yao2023react}
& \score{70.5}{1.6}
& \score{48.7}{4.4}
& \score{32.1}{2.1}
& \score{22.9}{3.6}
& \score{21.3}{2.3}
& \score{52.8}{2.4}
& \score{41.4}{1.3}
& \stepval{34.1}
\\

+ AWM~\citep{wang2025agent}
& \score{67.6}{3.3}
& \score{48.7}{4.4}
& \score{49.4}{2.2}
& \score{35.4}{3.6}
& \score{32.0}{4.0}
& \score{40.3}{2.4}
& \score{45.6}{1.8}
& \stepval{35.1}
\\

+ RBank~\citep{ouyang2026reasoningbank}
& \score{83.8}{1.6}
& \score{41.0}{4.4}
& \score{34.6}{2.2}
& \score{18.8}{6.3}
& \score{41.3}{2.3}
& \score{41.7}{4.2}
& \score{43.5}{2.2}
& \stepval{33.9}
\\

+ Trace2Skill~\citep{ni2026trace2skilldistilltrajectorylocallessons}
& \score{65.7}{2.9}
& \score{38.5}{0.0}
& \score{55.6}{3.7}
& \score{25.0}{6.2}
& \score{33.3}{2.3}
& \score{54.2}{0.0}
& \score{45.4}{1.7}
& \stepval{34.3}
\\

+ SkillTTA~\citep{wang2026skillsflytesttimeadaptive}
& \score{81.0}{1.7}
& \score{30.8}{7.7}
& \score{67.9}{4.3}
& \score{29.2}{3.6}
& \bestscore{54.7}{2.3}
& \score{43.1}{2.4}
& \score{51.1}{2.8}
& \stepval{30.4}
\\

+ \textbf{\textsc{ExpVoyager}} (Ours)
& \bestscore{88.6}{2.9}
& \bestscore{74.4}{4.4}
& \bestscore{72.9}{4.3}
& \bestscore{47.9}{3.6}
& \score{42.7}{2.3}
& \bestscore{55.6}{2.4}
& \bestscore{63.7}{2.2}
& \beststep{25.9}
\\

\midrule


\rowcolor[gray]{0.92}
\multicolumn{9}{c}{%
  \rule{0pt}{2.25ex}\itshape
  Executor $\theta$: Gemma4-31B
}
\\

Base (ReAct)~\citep{yao2023react}
& \score{91.4}{2.9}
& \score{56.4}{4.5}
& \score{39.5}{2.1}
& \score{52.1}{3.6}
& \score{53.3}{2.3}
& \score{30.6}{2.4}
& \score{53.9}{0.6}
& \stepval{29.2}
\\

+ AWM~\citep{wang2025agent}
& \score{87.6}{1.7}
& \score{59.0}{4.4}
& \score{54.3}{4.3}
& \bestscore{54.2}{3.6}
& \score{50.7}{2.3}
& \score{36.1}{2.4}
& \score{57.0}{0.9}
& \stepval{27.5}
\\

+ RBank~\citep{ouyang2026reasoningbank}
& \score{92.4}{1.7}
& \score{64.1}{4.4}
& \score{51.9}{3.7}
& \score{52.1}{7.2}
& \score{50.7}{4.6}
& \score{36.1}{6.4}
& \score{57.9}{1.5}
& \stepval{28.8}
\\

+ Trace2Skill~\citep{ni2026trace2skilldistilltrajectorylocallessons}
& \score{86.7}{4.4}
& \score{53.8}{7.7}
& \score{50.6}{2.1}
& \score{47.9}{9.5}
& \score{54.7}{4.6}
& \score{37.5}{4.2}
& \score{55.2}{2.4}
& \stepval{25.4}
\\

+ SkillTTA~\citep{wang2026skillsflytesttimeadaptive}
& \score{92.4}{1.7}
& \score{76.9}{7.7}
& \score{60.5}{4.3}
& \bestscore{54.2}{7.2}
& \score{50.7}{4.6}
& \score{40.3}{4.8}
& \score{62.5}{1.6}
& \stepval{24.7}
\\

+ \textbf{\textsc{ExpVoyager}} (Ours)
& \bestscore{94.3}{2.9}
& \bestscore{84.6}{7.7}
& \bestscore{82.7}{2.1}
& \bestscore{54.2}{3.6}
& \bestscore{57.3}{2.3}
& \bestscore{44.4}{2.4}
& \bestscore{69.6}{0.1}
& \beststep{22.0}
\\

\bottomrule
\end{tabular}%
}

\end{table*}
\begin{table*}[t]
\centering
\caption{
Detailed results on the ALFWorld and ScienceWorld in Figure~\ref{fig:budget}.
Tokens denote End-to-end tokens per task across the full task-time pipeline.
B01--B06 index budget groups, and $R$ is the number of
\textsc{ExpVoyager} navigation rounds.
Baselines use iterative retrieval in B01--B06.
Following each works original setup, one-shot settings are top-3 for AWM and SkillTTA and top-1 for ReasoningBank.
Bold marks the highest SR within each budget group
for each benchmark.
}
\label{tab:extended_budget_alfworld}
\small
\setlength{\tabcolsep}{4pt}
\renewcommand{\arraystretch}{1.0}
\setlength{\aboverulesep}{0.3ex}
\setlength{\belowrulesep}{0.4ex}
\resizebox{\linewidth}{!}{%
\begin{tabular}{@{}l*{4}{rr}@{}}
\toprule
\textbf{Budget ($R$)}
& \multicolumn{2}{c}{\textbf{\textsc{ExpVoyager}}}
& \multicolumn{2}{c}{\textbf{AWM}}
& \multicolumn{2}{c}{\textbf{ReasoningBank}}
& \multicolumn{2}{c}{\textbf{SkillTTA}} \\
\cmidrule(lr){2-3}\cmidrule(lr){4-5}
\cmidrule(lr){6-7}\cmidrule(lr){8-9}
& \textbf{Tokens} $\downarrow$ & \textbf{SR} $\uparrow$
& \textbf{Tokens} $\downarrow$ & \textbf{SR} $\uparrow$
& \textbf{Tokens} $\downarrow$ & \textbf{SR} $\uparrow$
& \textbf{Tokens} $\downarrow$ & \textbf{SR} $\uparrow$ \\
\midrule
\rowcolor{gray!12}
\multicolumn{9}{@{}l@{}}{\textbf{ALFWorld}\quad No-skill: 51,880.50 tokens/task; SR 41.4\%.} \\
\midrule
One-shot & -- & -- & 110,392.94 & 45.6 & 94,361.06 & 43.5 & 123,427.34 & 51.1 \\
\midrule
B01 ($R=5$) & 159,389.51 & \textbf{60.4} & 156,245.11 & 47.9 & 161,216.50 & 47.0 & 162,389.29 & 53.1 \\
B02 ($R=10$) & 239,787.24 & \textbf{62.9} & 244,231.09 & 48.8 & 233,414.12 & 45.5 & 239,236.17 & 51.9 \\
B03 ($R=15$) & 315,366.21 & \textbf{63.4} & 282,331.08 & 47.2 & 300,416.35 & 46.0 & 313,346.31 & 48.9 \\
B04 ($R=20$) & 385,314.17 & \textbf{63.7} & 378,191.30 & 45.9 & 371,782.17 & 43.6 & 405,290.17 & 49.1 \\
B05 ($R=25$) & 439,619.44 & \textbf{62.0} & 448,131.31 & 45.9 & 440,619.44 & 43.1 & 439,009.04 & 47.7 \\
B06 ($R=30$) & 480,549.49 & \textbf{58.8} & 491,139.12 & 45.0 & 480,117.07 & 42.9 & 478,132.18 & 46.5 \\
\midrule
\rowcolor{gray!12}
\multicolumn{9}{@{}l@{}}{\textbf{ScienceWorld}\quad No-skill: 40,722.28 tokens/task; SR 27.0\%.} \\
\midrule
One-shot & -- & -- & 45,629.32 & 32.6 & 36,728.71 & 29.6 & 69,037.24 & 29.6 \\
\midrule
B01 ($R=5$) & 131,518.27 & \textbf{35.9} & 135,123.11 & 33.7 & 123,118.30 & 31.9 & 131,238.36 & 28.7 \\
B02 ($R=10$) & 216,308.67 & \textbf{34.7} & 226,174.18 & 33.2 & 226,311.35 & 29.6 & 208,909.92 & 30.8 \\
B03 ($R=15$) & 286,391.10 & \textbf{36.6} & 279,311.10 & 31.7 & 282,121.41 & 28.7 & 265,191.10 & 29.2 \\
B04 ($R=20$) & 363,279.73 & \textbf{37.4} & 369,239.24 & 31.0 & 341,272.13 & 29.4 & 367,991.21 & 28.7 \\
B05 ($R=25$) & 419,819.22 & \textbf{37.7} & 418,817.12 & 30.1 & 421,123.15 & 28.5 & 439,111.42 & 27.3 \\
B06 ($R=30$) & 494,121.38 & \textbf{35.3} & 487,872.37 & 29.6 & 492,998.98 & 28.6 & 483,331.17 & 26.8 \\
\bottomrule
\end{tabular}%
}
\end{table*}
\begin{table}[t]
\centering
\caption{
Comparison of standalone \textsc{ExpVoyager} and its combinations with
pre-constructed skills, corresponding to the lower panels of
Figure~\ref{fig:synergy}.
Performance denotes success rate (\%) on ALFWorld and score on WebShop,
with higher values indicating better performance.
E2E tokens are reported per task and cover the full task-time pipeline,
including navigation and execution.
Gray rows denote combinations with pre-constructed skills.
Bold indicates improvements over standalone \textsc{ExpVoyager}
within each benchmark: higher performance or lower token cost.
}
\label{tab:extended_synergy}
\small
\setlength{\tabcolsep}{6pt}
\renewcommand{\arraystretch}{1.1}
\begin{tabular}{@{}llrr@{}}
\toprule
\textbf{Benchmark} & \textbf{Method}
& \textbf{SR / Score} $\uparrow$ & \textbf{E2E tokens/task} $\downarrow$ \\
\midrule
ALFWorld & \textsc{ExpVoyager} (alone) & 63.7 & 385,314 \\
\midrule
\rowcolor{gray!12}
& AWM + \textsc{ExpVoyager} & \textbf{65.3} & \textbf{320,103} \\
\rowcolor{gray!12}
& ReasoningBank + \textsc{ExpVoyager} & \textbf{64.9} & \textbf{298,582} \\
\rowcolor{gray!12}
& Trace2Skill + \textsc{ExpVoyager} & 59.6 & \textbf{293,893} \\
\midrule
WebShop & \textsc{ExpVoyager} (alone) & 61.2 & 390,997 \\
\midrule
\rowcolor{gray!12}
& AWM + \textsc{ExpVoyager} & \textbf{64.24} & \textbf{329,760} \\
\rowcolor{gray!12}
& ReasoningBank + \textsc{ExpVoyager} & \textbf{62.7} & \textbf{317,538} \\
\rowcolor{gray!12}
& Trace2Skill + \textsc{ExpVoyager} & 60.95 & \textbf{311,924} \\
\bottomrule
\end{tabular}
\end{table}

\section{Additional Results and Analysis}
\subsection{Detailed Results}

We provide detailed results for three experiments reported in the main text: (1) the ALFWorld subtask-wise breakdown of the main results in Table~\ref{tab:main_results}, with detailed results  in Table~\ref{tab:alfworld}; (2) the experience-access budget comparison in Figure~\ref{fig:budget}, with detailed results in Table~\ref{tab:extended_budget_alfworld}; and (3) the synergy between \textsc{ExpVoyager} and pre-constructed skills in Figure~\ref{fig:synergy}, with detailed results in Table~\ref{tab:extended_synergy}.

\subsection{Case Study}
\label{case_study}

To better understand how \textsc{ExpVoyager} navigates past experience and synthesizes target-conditioned skills, we conduct a qualitative analysis of successful and failed target executions across ALFWorld, WebShop, and ScienceWorld.
The \textit{cherry-picked} \textbf{success cases} provide a closer look at the navigation process , showing how the curator moves across experience views, follows evolving knowledge needs, and turns the observations into skill for the target task.
The \textit{lemon-picked} \textbf{failure cases} focus on the remaining challenges of experience navigation, examining where the navigation process can still fail to produce effective guidance for the target task.
Each figure presents the target task, example navigation rounds, the corresponding navigation observations, and updated navigation state.

\subsubsection{Success Cases}

\paragraph{Distinguishing Source-Specific Observations from Target-Relevant Knowledge.}
Figure~\ref{fig:case_alf_success_1} illustrates how the curator distinguishes a source-specific object location from a procedure that can transfer to the current target.
The ALFWorld task requires \textit{placing a newspaper on a sofa}.
An initially accessed source trajectory obtains the \textit{newspaper from a coffee table}, but the target's initial observation lists \textit{no coffee table}.
This discrepancy redirects navigation from the placement action to alternative newspaper locations.
Using \texttt{search\_exp} in the step-level view, the curator locates a record in which \textit{a newspaper is obtained from an armchair}, and \texttt{inspect\_traj} expands this record into the surrounding search and placement sequence.
The curator interprets these navigation observations as support for a reusable search-and-placement procedure, rather than evidence that the newspaper occupies the same location in the target environment.
The synthesized skill retains an execution-time check of the expected source and recovery guidance to broaden the search when it is empty or unavailable.
The frozen executor ultimately finds \textit{a newspaper on a side table} and completes the task.

\paragraph{Connecting Actions to Their Observed Results.}
Figure~\ref{fig:case_alf_success_2} demonstrates how the curator uses local execution records and their surrounding context to establish a required action and the observations needed to check its effect.
The ALFWorld task requires \textit{cleaning a soap bar and placing it in a toilet}.
Through \texttt{inspect\_traj}, the curator examines observations of \textit{the same soap bar before and after cleaning}, connecting the action to the observed change in object state.
A further \texttt{search\_exp} call over the reasoning field in the step-level view accesses a record explaining that \textit{the agent already holds the soap bar} and should therefore proceed with cleaning even though \textit{the sink appears empty}.
Together, these navigation observations clarify the condition under which the action can proceed and the result that should be checked afterward.
The synthesized skill requires cleaning before placement and retains verification of the cleaning result as an execution-time check.
With this guidance, the executor completes the task, whereas the base agent without skills never issues a \textit{cleaning action}.

\paragraph{Preserving Action Ordering through Cross-Step Context.}
Figure~\ref{fig:case_alf_success_3} shows how expanding local records into broader trajectory context helps the curator identify the ordering required for the target task.
The ALFWorld task requires \textit{cooling a plate and placing it in a cabinet}.
An initial source trajectory provides the cabinet-placement action, while a later \texttt{inspect\_traj} call exposes a sequence in which \textit{an early placement is followed by cooling and another placement}.
A subsequent \texttt{search\_exp} call in the step-level view supplies a concrete cooling command.
This cross-step context distinguishes an intermediate action from the final placement that follows the required change in object state.
The curator interprets these navigation observations together to establish that cooling must precede the final placement.
The synthesized skill turns this procedural knowledge into an ordered sequence: \textit{acquire the plate, cool it with a refrigerator, and then place it in the cabinet}.
The frozen executor follows this ordering and completes the task.

\paragraph{Retaining Target Conditions as Execution-Time Checks.}
Figure~\ref{fig:case_web_success_1} illustrates how the curator interprets past decisions to identify conditions that still require verification during target execution.
The WebShop task requires \textit{black jeans in a specified size}, but the relevant attributes may appear either as selectable options or as product text.
Using \texttt{search\_exp} over recorded reasoning in the step-level view, the curator accesses a record warning that \textit{an immediate purchase could leave the product in an unintended default variant}.
Through \texttt{inspect\_traj}, the curator also examines \textit{a purchase made without selecting the requested options after opening a new product}.
These navigation observations support a procedural rule that depends on how an attribute is presented: selectable options require explicit interaction, whereas descriptive attributes require checking the product text.
The synthesized skill retains these conditions as execution-time checks for the current product.
The frozen executor \textit{selects black and large before purchasing}, applying the procedure to the options available in the target environment.

\paragraph{Checking Target Conditions against Available Options.}
Figure~\ref{fig:case_web_success_2} demonstrates how procedural knowledge from past experience guides verification of the options available in the current environment.
The WebShop task requires \textit{grey memory-foam slippers}.
The curator uses \texttt{search\_exp} in the step-level view to access recorded reasoning about checking product details, then inspects a source trajectory in which \textit{the agent leaves an unsuitable candidate and selects an option on a replacement product}.
These navigation observations support execution-time checks of the requested attributes and recovery guidance for replacing a candidate that cannot satisfy them.
During target execution, the first product mentions \textit{``Black-Grey''} in its description, but its selectable colors are \textit{pink and lake blue}.
Guided by the synthesized skill, the frozen executor \textit{moves to another candidate, selects its grey option, and completes the purchase}.
The transferred knowledge is therefore a verification-and-recovery procedure, rather than an assumption that related wording in a product description establishes the availability of a requested option.

\paragraph{Identifying Required Conditions across Source Trajectories.}
Figure~\ref{fig:case_sci_success_1} shows how the curator combines navigation observations from different source trajectories to identify a condition needed for the current target.
The ScienceWorld task requires \textit{making a peanut-butter sandwich}.
Through \texttt{inspect\_traj}, the curator examines a failed sandwich attempt in which \textit{acquiring the ingredients and then mixing an ingredient produces no result}.
A \texttt{search\_exp} call over the reasoning field in the step-level view accesses a record from a mixed-nuts task that connects mixing to \textit{ingredients already sharing a container}.
Further trajectory inspection clarifies that \textit{the sandwich ingredients had remained in inventory rather than being transferred into a common container}.
Interpreting these observations together, the curator identifies the shared-container requirement as procedural knowledge that can transfer across recipes.
The synthesized skill turns this condition into execution guidance: \textit{transfer the ingredients into a common container before mixing its contents}.
The frozen executor \textit{transfers the ingredients into a cup and successfully produces the sandwich}.

\paragraph{Interpreting Recorded Reasoning for the Current Target.}
Figure~\ref{fig:case_sci_success_2} illustrates how recorded reasoning helps the curator interpret why an action was taken and what part of that decision can transfer to the current target.
The ScienceWorld task requires \textit{identifying the animal with the longest lifespan and then the one with the shortest lifespan}.
Using \texttt{search\_exp} over the reasoning field in the step-level view, the curator accesses a record explaining that \textit{examining another animal serves to compare the available candidates}, even though the associated observation contains no numerical lifespan information.
Through \texttt{inspect\_traj}, the curator also examines a source sequence that establishes the required order of the two focus actions, while additional navigation observations expose different candidate sets.
The curator separates these source-specific candidates from the reusable procedure for comparing the animals present and applying the required selection order.
The synthesized skill guides the executor to identify the available animals and \textit{focus on the longest-lived candidate before the shortest-lived one}.
The frozen executor completes the task by focusing on \textit{the tortoise egg and then the dragonfly}.

\subsubsection{Failure Cases}

\paragraph{Losing Action Ordering during Skill Synthesis.}
Figure~\ref{fig:case_alf_fail_1} illustrates a failure to retain ordering constraints from navigation observations in the final skill.
The ALFWorld task requires \textit{placing two remote controls on an ottoman}.
Through \texttt{inspect\_traj}, the curator examines a source trajectory that handles two objects sequentially: \textit{acquire the first, place it, and then repeat for the second}.
A further \texttt{search\_exp} match in the step-level view explicitly records the reasoning that \textit{the currently held remote should be placed before searching for another}.
Despite these supporting records, the synthesized skill requires \textit{acquiring both remotes before beginning placement}.
The frozen executor consequently \textit{continues searching while holding one remote and never issues a placement action}.
This case suggests that, even when the relevant ordering constraints are successfully identified during navigation, experience navigation can still fail to preserve them when multiple observations are consolidated into a single executable skill.

\paragraph{Failing to Recheck Target Conditions during Execution.}
Figure~\ref{fig:case_web_fail_1} presents a failure involving conditions that must be checked again after the executor revisits a product.
The WebShop task requires \textit{a purple, XX-Large, officially licensed Batman shirt}.
During navigation, the curator accesses records concerning explicit option selection and recorded reasoning about uncertainty when option clicks leave observations visually similar.
During target execution, the frozen executor \textit{selects the requested color and size but subsequently leaves and reopens the product}.
After its final return, it \textit{purchases without selecting those options again}, and the purchase record contains \textit{an empty options map}.
Thus, earlier option-selection actions do not establish that the required selections remain active at the moment of purchase.
This case reveals a remaining challenge in adapting knowledge as the execution state changes. In   particular, more execution-state-aware navigation and knowledge management could help determine when previously satisfied conditions should be checked again before consequential actions such as purchase.

\paragraph{Translating Required Conditions into Execution Guidance.}
Figure~\ref{fig:case_sci_fail_1} demonstrates a gap between procedural knowledge established during navigation and the actions specified in the synthesized skill.
The ScienceWorld task requires \textit{making a peanut-butter-and-jam sandwich}.
By examining failed recipe trajectories and a successful mixing example, the curator identifies that \textit{the ingredients must share a container before mixing} and retains this condition in the Navigation State.
However, the final skill incorrectly proposes a mixing command, such as \texttt{mix kitchen}, as a way to \textit{transfer the ingredients into a container}.
The frozen executor \textit{gathers the ingredients but repeatedly attempts to mix the kitchen without first transferring them into a common container}.
In contrast to the successful sandwich case in Figure~\ref{fig:case_sci_success_1}, the required condition is identified, but the final skill does not specify the actions needed to establish it.
This case shows that identifying the correct precondition is not sufficient when the skill cannot determine the valid action sequence needed to satisfy it. A remaining challenge is therefore to extend navigation beyond identifying the required precondition toward finding the concrete actions needed to establish that state in the target environment.

\clearpage


\newcommand{\casefigure}[3]{%
\begin{figure*}[t]
    \centering
    \includegraphics[width=\textwidth]{#1}
    \caption{#2}
    \label{#3}
    \vspace{-2mm}
\end{figure*}
}


\casefigure
{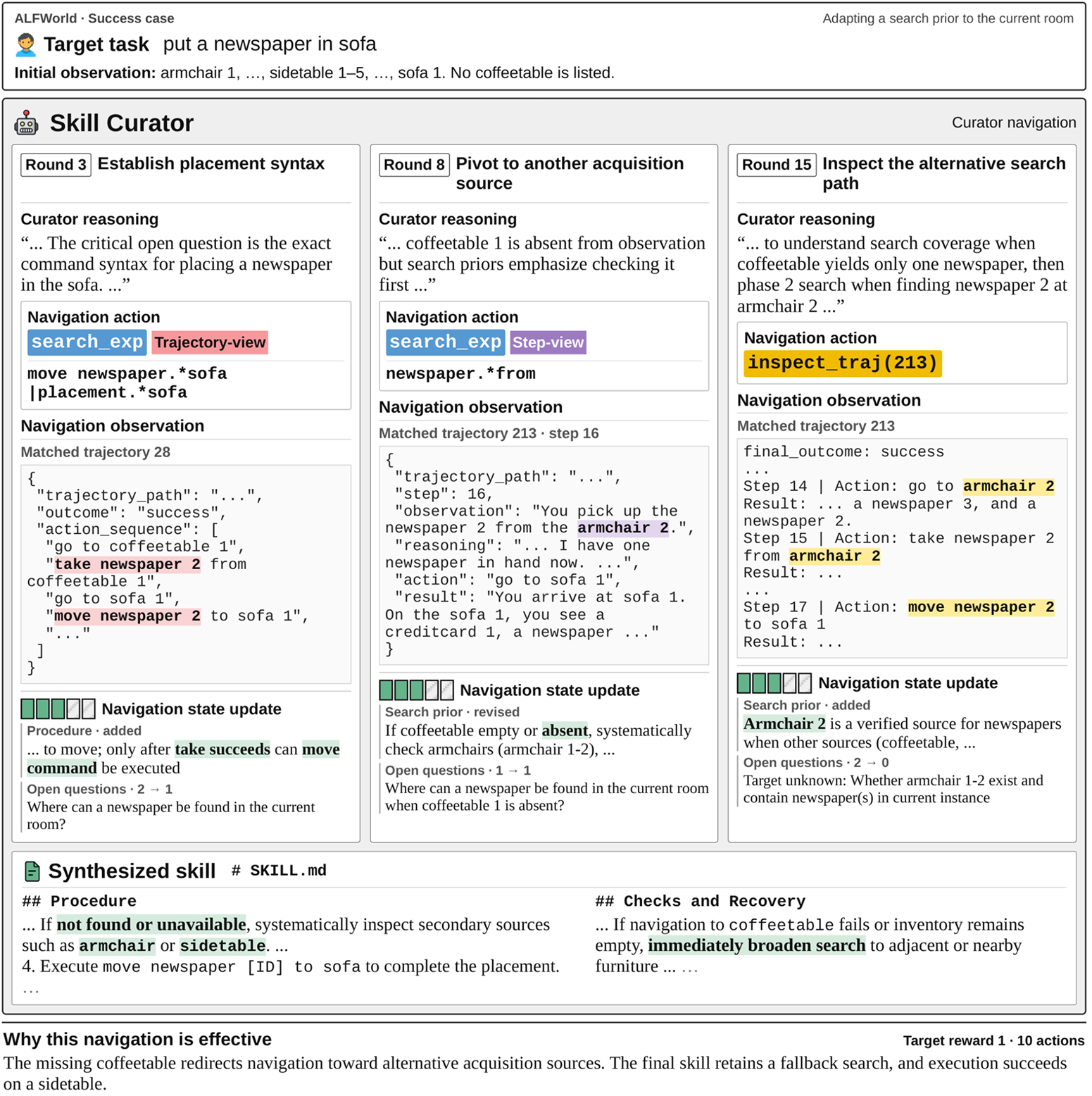}
{\textbf{Successful example (ALFWorld): distinguishing source-specific observations from target-relevant knowledge.}
The absence of a coffee table in the target's initial observation redirects navigation from the placement action to alternative newspaper locations.
The curator interprets a step-level \texttt{search\_exp} match together with the cross-step context returned by \texttt{inspect\_traj}, distinguishing a reusable search-and-placement procedure from a source-specific object location.
The final skill retains a location check and fallback search as execution guidance.}
{fig:case_alf_success_1}

\casefigure
{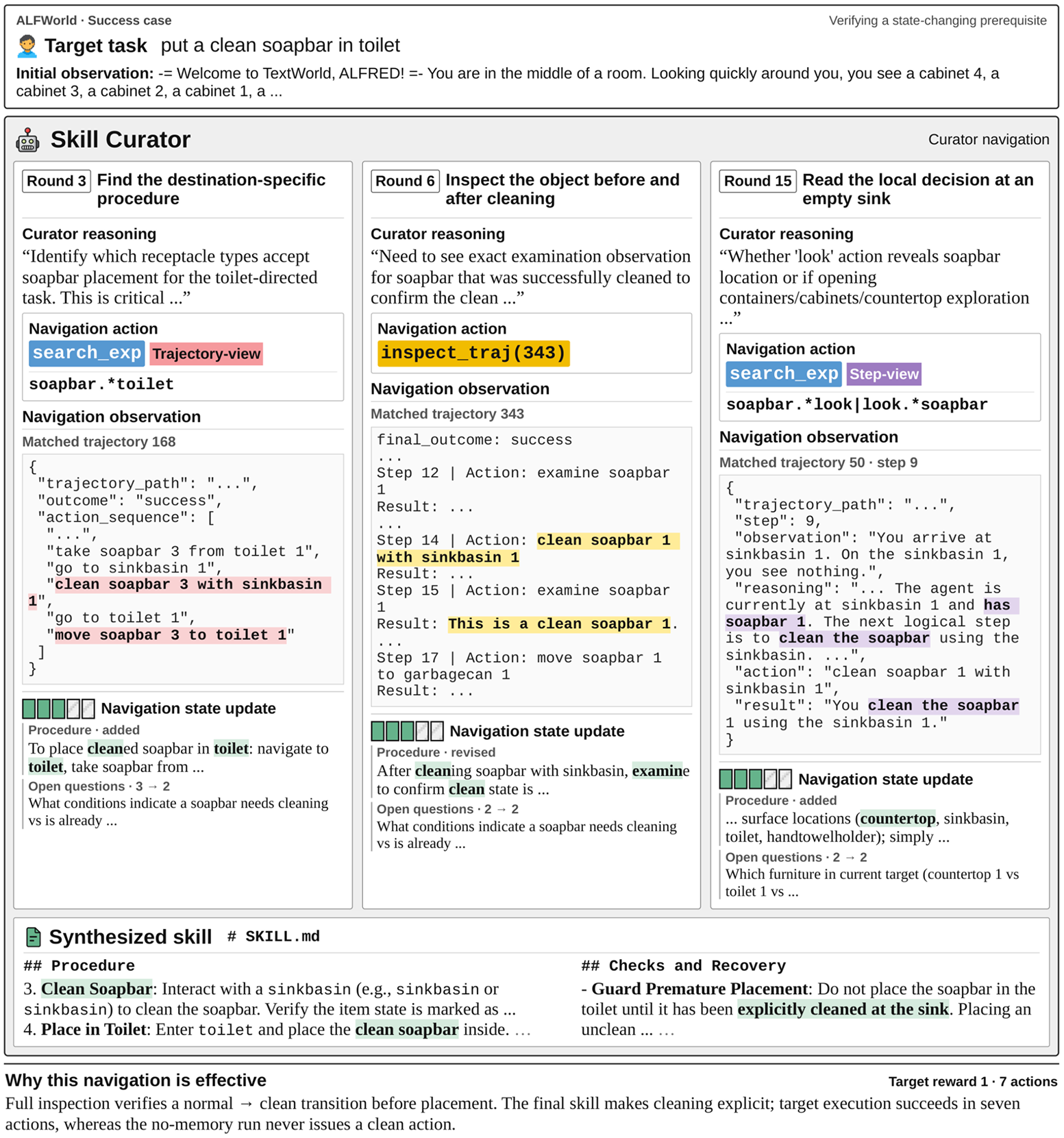}
{\textbf{Successful example (ALFWorld): connecting actions to their observed results.}
Through \texttt{inspect\_traj}, the curator connects observations of the same soap bar before and after cleaning, while recorded reasoning in a step-level match clarifies when the cleaning action can proceed.
The synthesized skill preserves cleaning before placement and retains verification of its result as an execution-time check.
The frozen executor follows this ordering and completes the task.}
{fig:case_alf_success_2}

\casefigure
{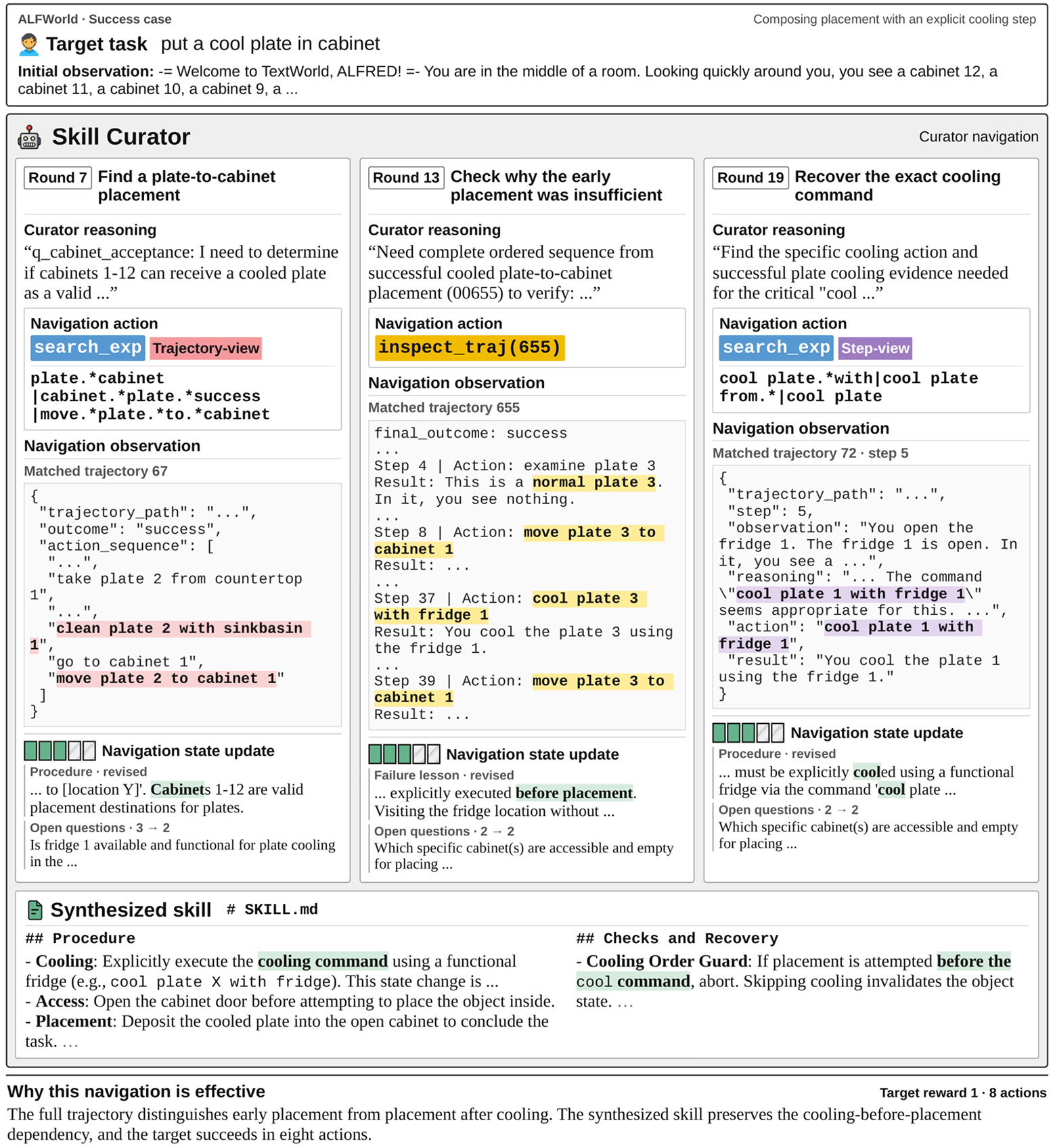}
{\textbf{Successful example (ALFWorld): preserving action ordering through cross-step context.}
The cross-step context returned by \texttt{inspect\_traj} distinguishes an early placement from the final placement after cooling.
A step-level \texttt{search\_exp} match supplies a concrete cooling command.
The curator interprets these navigation observations together and retains the required cooling-before-placement ordering in the synthesized skill.}
{fig:case_alf_success_3}

\casefigure
{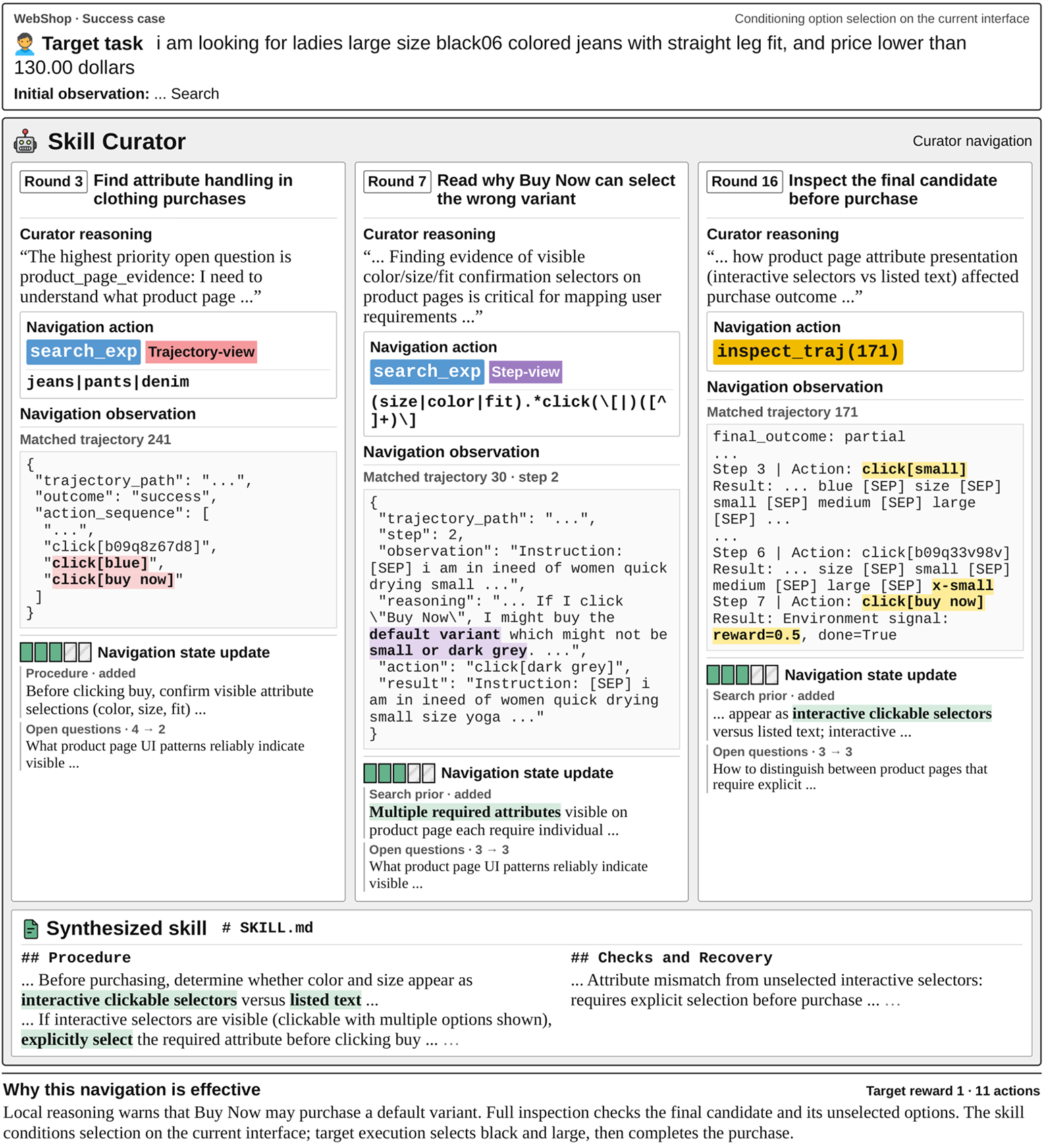}
{\textbf{Successful example (WebShop): retaining target conditions as execution-time checks.}
Recorded reasoning accessed through \texttt{search\_exp} warns against purchasing a default variant, while \texttt{inspect\_traj} exposes a purchase without option selection after opening a new product.
The synthesized skill distinguishes selectable options from descriptive attributes and retains the corresponding execution-time checks.
The frozen executor selects the requested color and size before purchasing.}
{fig:case_web_success_1}

\casefigure
{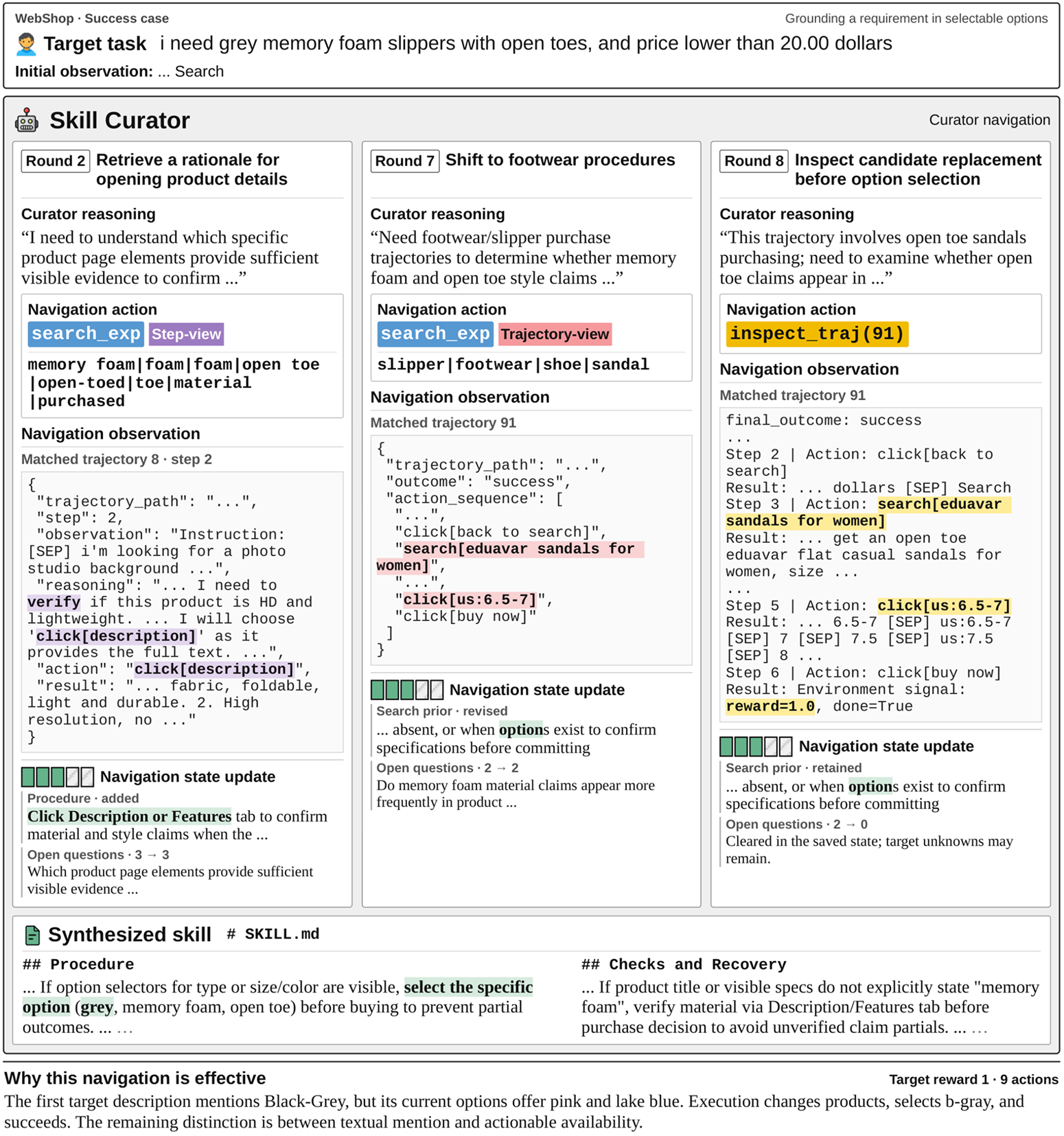}
{\textbf{Successful example (WebShop): checking target conditions against available options.}
Navigation observations support execution-time checks of required attributes and recovery guidance for replacing an unsuitable product.
During target execution, grey appears in the first product's description but is unavailable among its selectable colors.
The frozen executor switches to another candidate and selects its grey option before purchasing.}
{fig:case_web_success_2}

\casefigure
{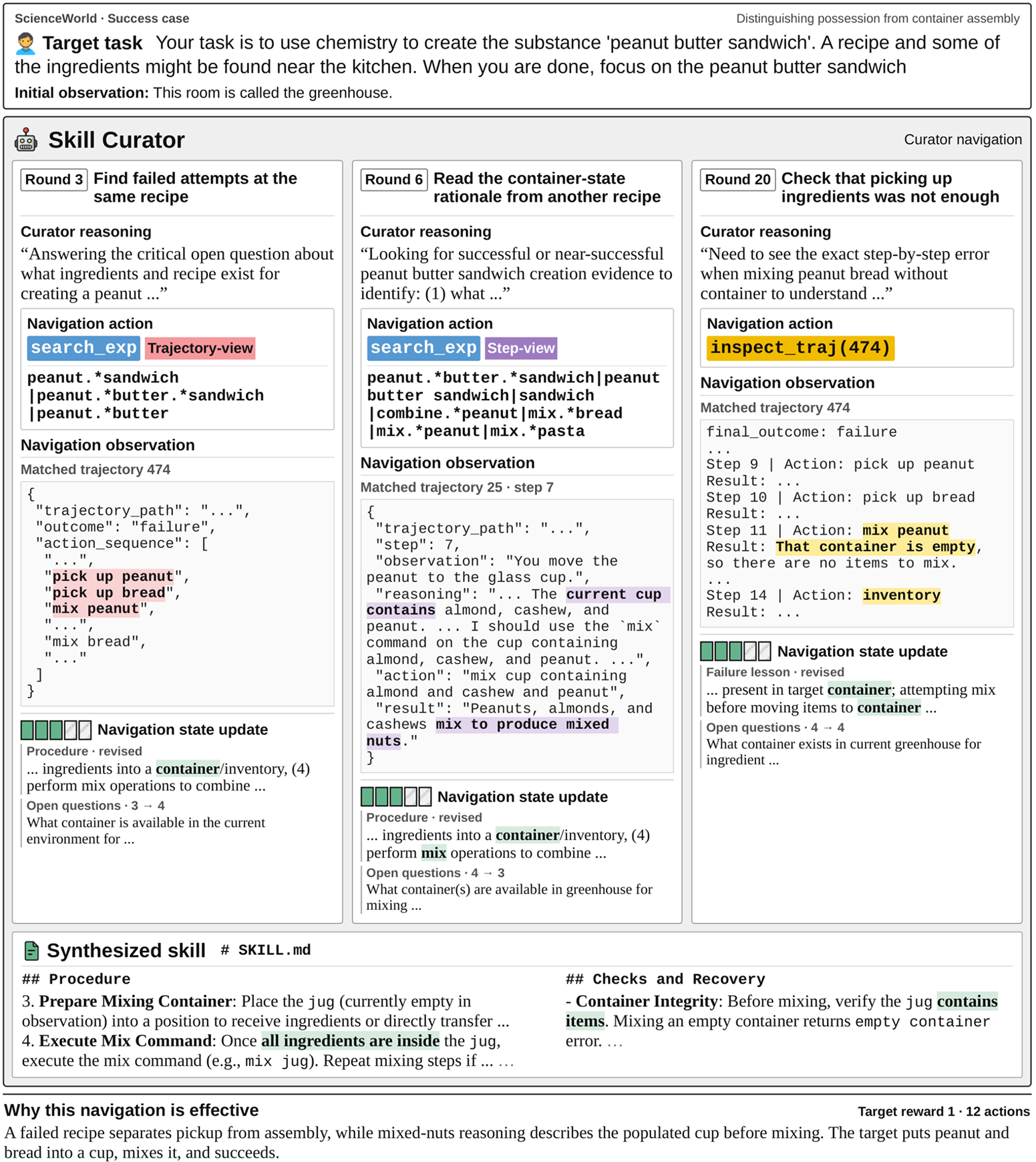}
{\textbf{Successful example (ScienceWorld): identifying required conditions across source trajectories.}
The curator interprets a failed sandwich trajectory together with recorded reasoning from another recipe to identify that the ingredients must share a container before mixing.
The synthesized skill turns this procedural knowledge into explicit ingredient-transfer and mixing actions.
The frozen executor moves the ingredients into a cup and successfully produces the sandwich.}
{fig:case_sci_success_1}

\casefigure
{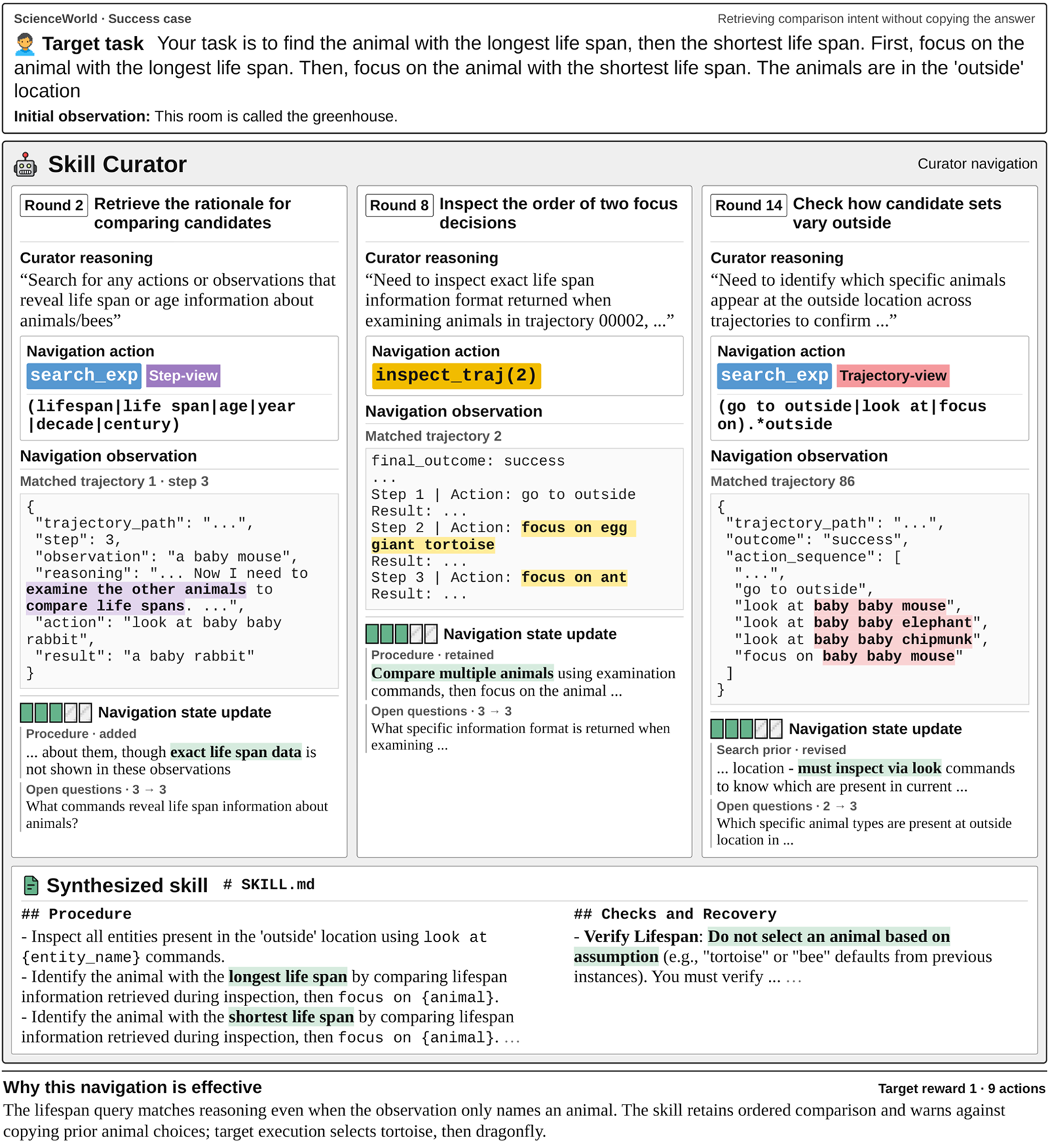}
{\textbf{Successful example (ScienceWorld): interpreting recorded reasoning for the current target.}
A \texttt{search\_exp} match over the reasoning field in the step-level view exposes the comparison purpose behind examining additional animals, while \texttt{inspect\_traj} establishes the required focus-action order.
The curator separates source-specific candidates from the reusable comparison procedure.
The synthesized skill guides the frozen executor to apply this procedure to the animals available in the current environment.}
{fig:case_sci_success_2}


\casefigure
{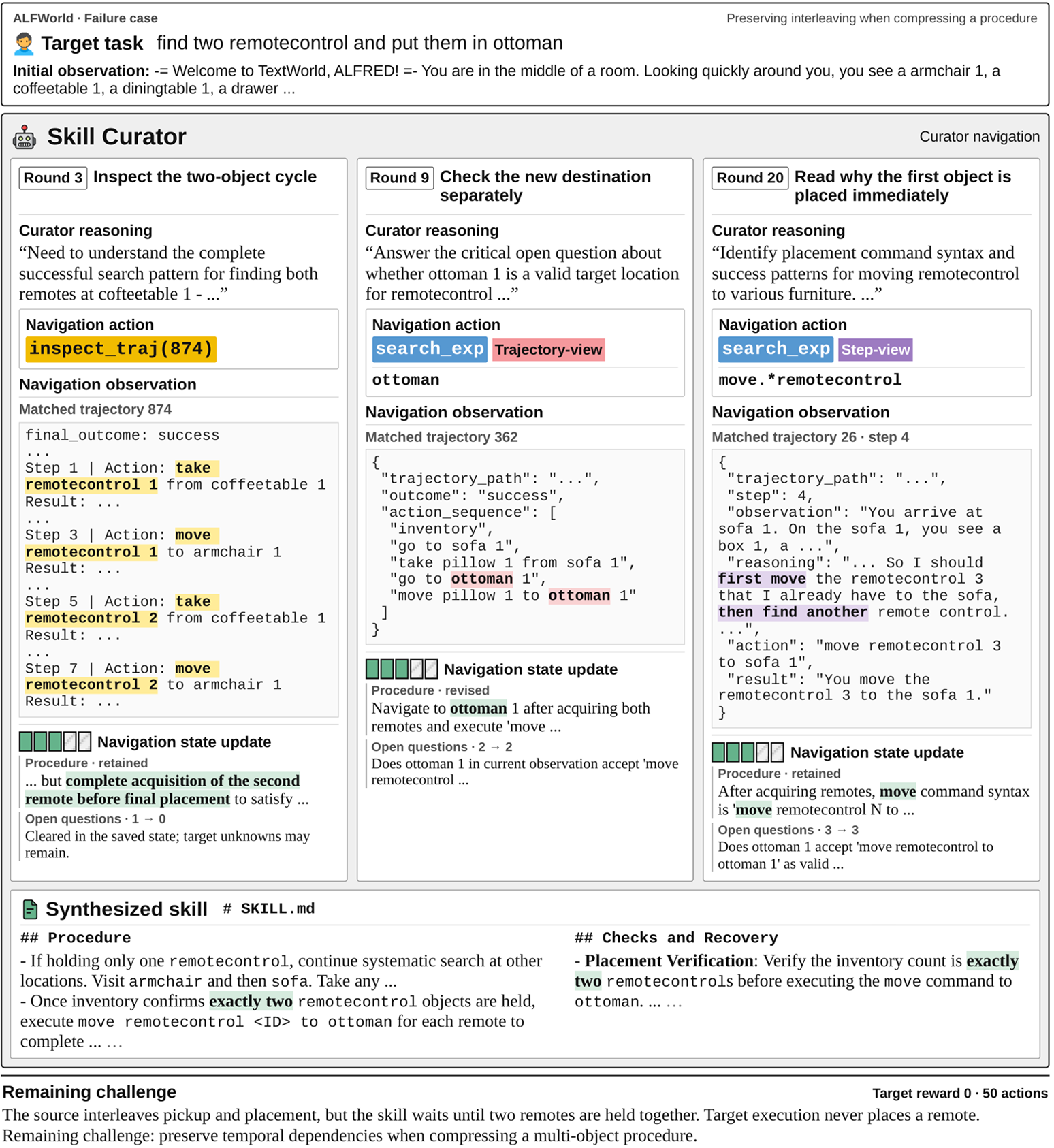}
{\textbf{Failure example (ALFWorld): losing action ordering during skill synthesis.}
Navigation observations show that acquisition and placement should alternate, but the final skill requires both remote controls to be acquired before placement.
The frozen executor continues searching while holding one remote and never issues a placement action.}
{fig:case_alf_fail_1}

\casefigure
{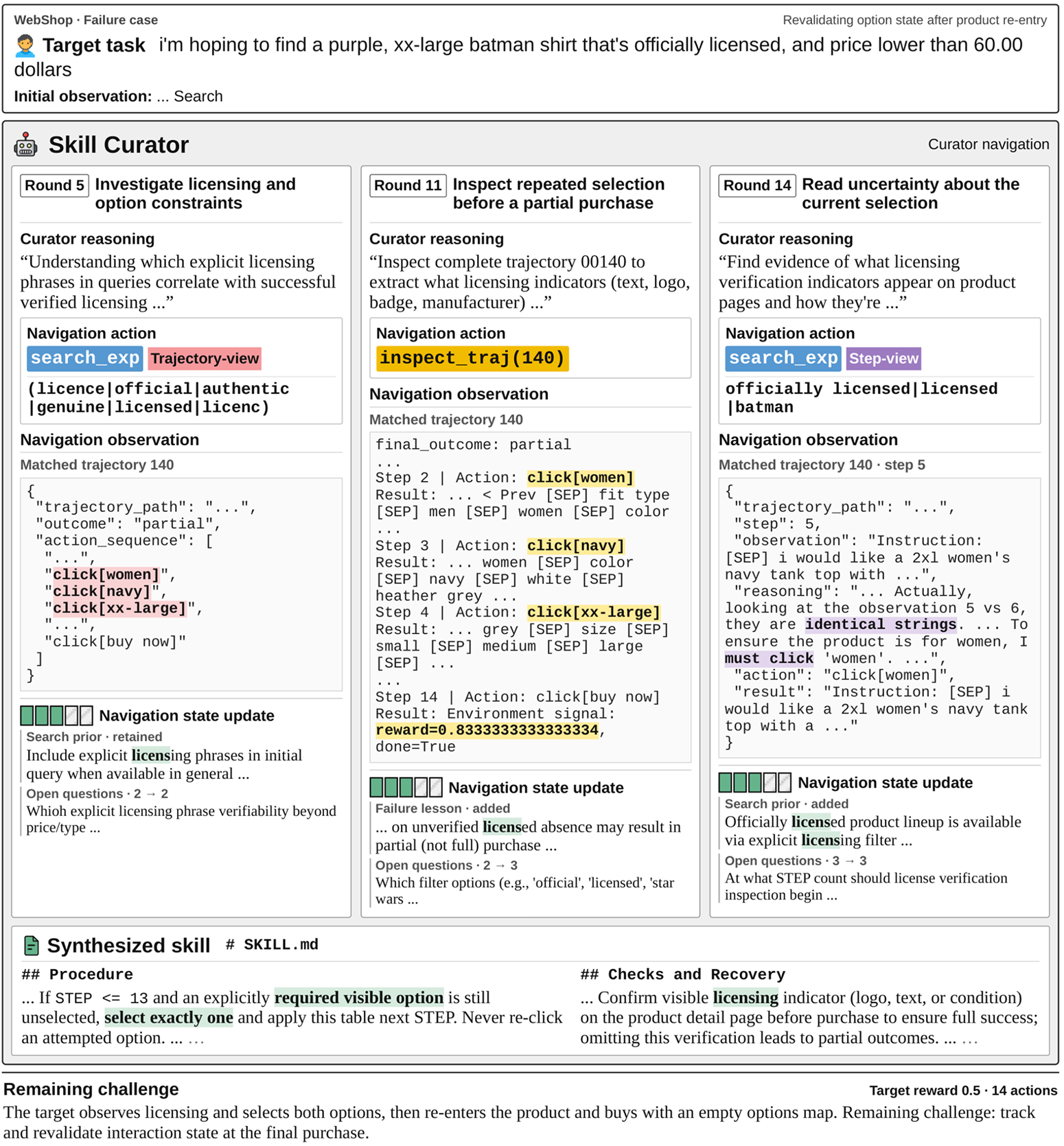}
{\textbf{Failure example (WebShop): failing to recheck target conditions during execution.}
The frozen executor selects the requested color and size but later purchases after reopening the product without reselecting those options.
The empty options map in the purchase record shows that earlier selections do not establish the required state at purchase time.}
{fig:case_web_fail_1}

\casefigure
{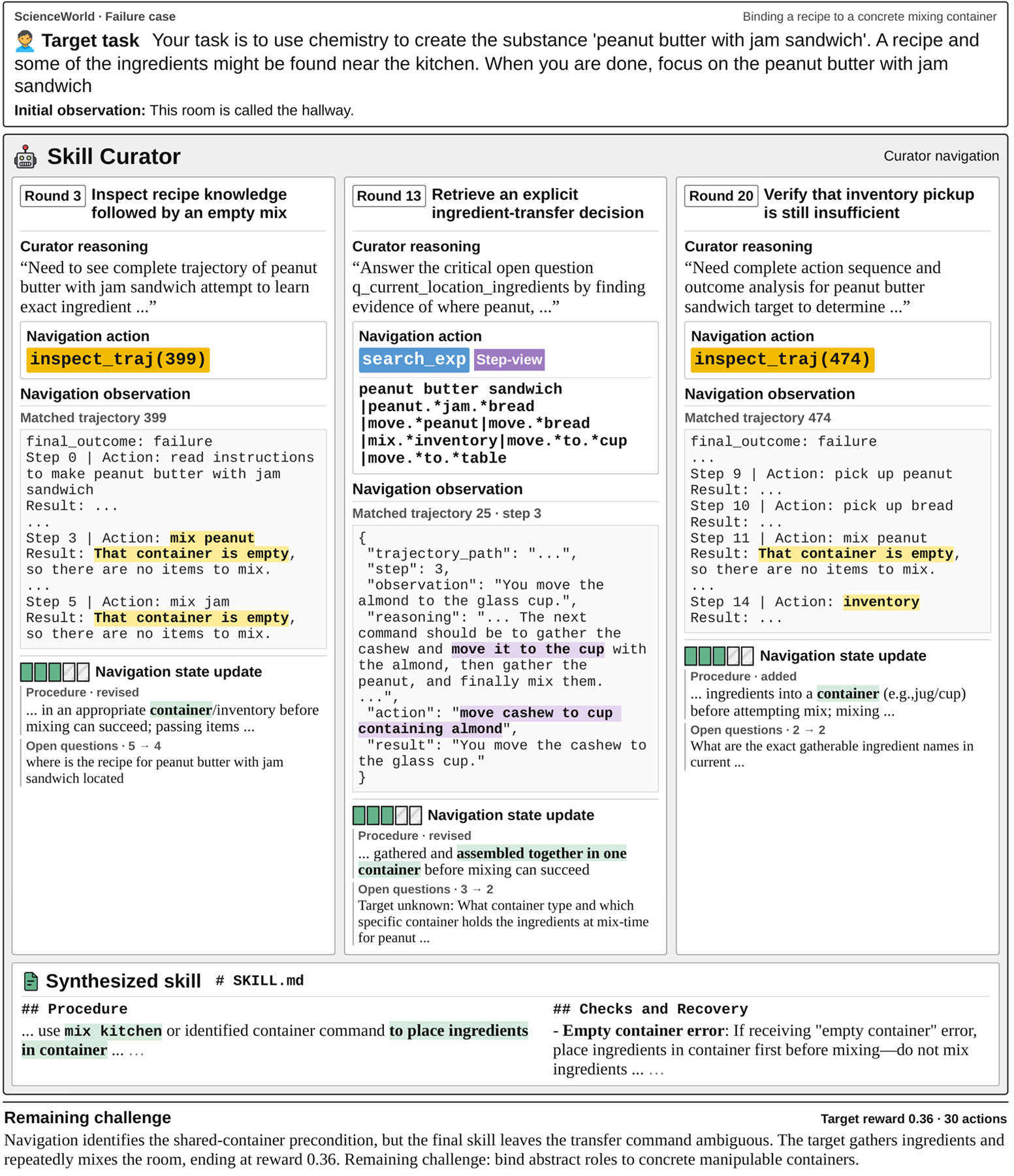}
{\textbf{Failure example (ScienceWorld): translating required conditions into execution guidance.}
The Navigation State retains the requirement that ingredients share a container, but the final skill incorrectly proposes \texttt{mix kitchen} as an ingredient-transfer action.
The frozen executor repeatedly attempts to mix without first moving the ingredients into a common container.
The required condition is identified during navigation but is not translated into actions that establish it.}
{fig:case_sci_fail_1}
\clearpage
\begingroup
\raggedbottom
\tcbset{evcuratorprompt/.style={
  enhanced, breakable, width=\linewidth,
  colback=black!2, colframe=black!50,
  colbacktitle=black!65, coltitle=white,
  fonttitle=\normalsize\bfseries, fontupper=\footnotesize\raggedright,
  boxrule=0.5pt, arc=2pt,
  left=8pt, right=8pt, top=6pt, bottom=6pt,
  toptitle=4pt, bottomtitle=4pt,
  before skip=0pt, after skip=0pt,
  before upper={\setlength{\parindent}{0pt}\setlength{\parskip}{0pt}}
}}

\par
\begingroup
\captionsetup{position=top,skip=0pt}
\captionof{table}{System prompt for Navigation Action Selection.}
\label{tab:navigation_action_selection_prompt}
\par\nopagebreak[4]\vspace{12pt}
\begin{tcolorbox}[evcuratorprompt,breakable=false,title={Navigation Action Selection: system message}]
You investigate an outcome-aware trajectory corpus using one role-separated Navigation State as the sole persistent memory of the investigation.

\smallskip
The goal and its requirements come only from the current target. The remaining State distinguishes established procedures, evidence-informed search priors, failure lessons, and open questions. Choose the highest-priority open question whose answer would most improve the final executable skill. Do not draft or revise a Working Skill during search.

\smallskip
The corpus exposes two evidence levels. trajectory\_index.jsonl contains one episode-level outcome judgment. Use it for high-level success causes, failure causes, recovery patterns, and reusable strategy. step\_index.jsonl contains raw local transitions. Use it for exact preconditions, executed actions, state changes, and environment feedback. There is no fixed order between the two indexes; choose the level needed by the current question.

\smallskip
Exact command, ordering, or state-change claims require a step row or linked chronological trajectory. Source-model reasoning explains intent or hypothesis; observation, executed action, and result establish what locally happened.

\smallskip
search\_exp returns complete raw rows rather than a model-authored summary. Treat returned rows as an unranked evidence set. Read an exact trajectory\_path when whole-episode ordering, consequences, recovery, or grounding is needed. If returned evidence does not answer the stated question, let the State updater reject it rather than extracting a convenient but unrelated pattern.

\smallskip
The current target and observation are authoritative. A source episode cannot establish the current instance's location, id, or state.

\smallskip
Before calling a tool, state the open question it should answer and why that answer matters to the final skill. Call exactly one search\_exp/inspect\_traj tool, or respond exactly with FINISH when the State contains enough grounded knowledge to synthesize a useful skill and another retrieval is unlikely to improve it. Do not write skill.md on a search turn.
\end{tcolorbox}
\endgroup
\clearpage
\endgroup
\clearpage
\begingroup
\raggedbottom
\tcbset{evcuratorprompt/.style={
  enhanced, breakable, width=\linewidth,
  colback=black!2, colframe=black!50,
  colbacktitle=black!65, coltitle=white,
  fonttitle=\normalsize\bfseries, fontupper=\footnotesize\raggedright,
  boxrule=0.5pt, arc=2pt,
  left=8pt, right=8pt, top=6pt, bottom=6pt,
  toptitle=4pt, bottomtitle=4pt,
  before skip=0pt, after skip=0pt,
  before upper={\setlength{\parindent}{0pt}\setlength{\parskip}{0pt}}
}}

\par
\begingroup
\captionsetup{position=top,skip=0pt}
\captionof{table}{User-message template for Navigation Action Selection after a corpus search. Before the first search, the active-evidence block is \texttt{Active evidence set:} followed by \texttt{(none; search the evidence level needed by the current question)}. The last-action feedback block is included only when feedback is available.}
\label{tab:navigation_action_selection_user}
\par\nopagebreak[4]\vspace{12pt}
\begin{tcolorbox}[evcuratorprompt,breakable=false,title={Navigation Action Selection: user message}]
\textbf{Target:}\par
\texttt{\{target\_task\}}

\smallskip
\textbf{Current observation:}\par
\texttt{\{current\_observation\}}

\smallskip
\textbf{Admissible commands:}\par
\texttt{\{admissible\_commands\}}

\smallskip
\textbf{Navigation State:}\par
\texttt{\{current\_state\}}

\smallskip
\textbf{Corpus evidence levels:}\par
\begin{itemize}[leftmargin=*,nosep,topsep=1pt]
\item \texttt{trajectory\_index.jsonl}: one row per source episode with \texttt{trajectory\_path}, \texttt{outcome}, and exact ordered \texttt{action\_sequence}. Search \texttt{action\_sequence} for multi-action ordering.
\item \texttt{step\_index.jsonl}: one row per raw transition with outcome context and exact local evidence (\texttt{observation}, \texttt{reasoning}, \texttt{action}, \texttt{result}).
\item \texttt{trajectory\_path}: the complete source episode, including its outcome judgement and chronological Action -\textgreater{} Result path. It remains the source of truth for whole-path grounding.
\end{itemize}

\smallskip
\textbf{Navigation rules:}\par
\begin{itemize}[leftmargin=*,nosep,topsep=1pt]
\item Search exactly one index and only the fields needed by the current Navigation State question; every match is returned as the complete stored row.
\item Compare multiple trajectory candidates through \texttt{outcome}, and \texttt{action\_sequence} when the question concerns procedure ordering, search coverage, failure, recovery, or completion.
\item Use a step match as complete local evidence. Read its exact \texttt{trajectory\_path} only when the whole path is needed.
\item Use \texttt{action\_sequence} to locate an observed action ordering, not as proof of effects; validate effects in step rows or the linked raw trajectory.
\item Choose the evidence level and fields from the current question; no trajectory-first or step-first order is mandatory.
\end{itemize}

\smallskip
\textbf{Active evidence set (latest corpus search):}\par
\textbf{Search query:}\par
\texttt{\{latest\_search\_query\}}\par
\textbf{Candidate inspection status:}\par
\texttt{\{candidate\_inspection\_status\}}\par
\textbf{Returned evidence rows:}\par
\texttt{\{returned\_evidence\_rows\}}

\smallskip
\textbf{Last action feedback:}\par
\texttt{\{navigation\_feedback\}}

\smallskip
Choose an open Navigation State question whose answer could most improve the final executable skill, taking critical items before high or optional items. First use an active candidate that can answer it; issue a new query only when the active set cannot. Do not repeat a query unless evidence has changed.

\smallskip
\textbf{For a tool call, provide arguments in this order:}\par
\begin{enumerate}[leftmargin=*,nosep,topsep=1pt]
\item \texttt{reason}: the Navigation State question to answer and why its answer matters to the final skill.
\item For search\_exp, choose the evidence-level \texttt{index}, a focused \texttt{query}, the fields to search, and a bounded result limit. For inspect\_traj, use an exact returned \texttt{trajectory\_path}.
\end{enumerate}

\smallskip
Respond exactly with FINISH only when the latest valid evidence is stored in the Navigation State, the State supports a useful skill, and no further corpus action is likely to improve it.
\end{tcolorbox}
\endgroup
\clearpage
\endgroup
\clearpage
\begingroup
\raggedbottom
\tcbset{evcuratorprompt/.style={
  enhanced, breakable, width=\linewidth,
  colback=black!2, colframe=black!50,
  colbacktitle=black!65, coltitle=white,
  fonttitle=\normalsize\bfseries, fontupper=\footnotesize\raggedright,
  boxrule=0.5pt, arc=2pt,
  left=8pt, right=8pt, top=6pt, bottom=6pt,
  toptitle=4pt, bottomtitle=4pt,
  before skip=0pt, after skip=0pt,
  before upper={\setlength{\parindent}{0pt}\setlength{\parskip}{0pt}}
}}

\par
\begingroup
\captionsetup{position=top,skip=0pt}
\captionof{table}{Native function-schema excerpt for \texttt{search\_exp}. Numeric bounds on \texttt{limit} are omitted.}
\label{tab:search_exp_tool_schema}
\par\nopagebreak[4]\vspace{12pt}
\begin{tcolorbox}[evcuratorprompt,breakable=false,title={Navigation tool: \texttt{search\_exp} schema}]
\begingroup
\ttfamily\footnotesize\raggedright
\setlength{\parindent}{0pt}
\setlength{\parskip}{0pt}
\noindent\hangindent=0em\hangafter=1\hspace*{0em}\{\par
\noindent\hangindent=1em\hangafter=1\hspace*{1em}"type": "function",\par
\noindent\hangindent=1em\hangafter=1\hspace*{1em}"function": \{\par
\noindent\hangindent=2em\hangafter=1\hspace*{2em}"name": "search\_exp",\par
\noindent\hangindent=2em\hangafter=1\hspace*{2em}"description": "Search one corpus JSONL index in selected evidence fields and return complete matching evidence rows. An exact returned trajectory\_path may be opened with inspect\_traj when the full episode is needed.",\par
\noindent\hangindent=2em\hangafter=1\hspace*{2em}"parameters": \{\par
\noindent\hangindent=3em\hangafter=1\hspace*{3em}"type": "object",\par
\noindent\hangindent=3em\hangafter=1\hspace*{3em}"properties": \{\par
\noindent\hangindent=4em\hangafter=1\hspace*{4em}"reason": \{"type": "string", "description": "Why this query should answer the active question"\},\par
\noindent\hangindent=4em\hangafter=1\hspace*{4em}"index": \{\par
\noindent\hangindent=5em\hangafter=1\hspace*{5em}"type": "string",\par
\noindent\hangindent=5em\hangafter=1\hspace*{5em}"enum": ["trajectory\_index.jsonl", "step\_index.jsonl"],\par
\noindent\hangindent=5em\hangafter=1\hspace*{5em}"description": "Choose one evidence index and use only its allowed fields. trajectory\_index.jsonl allowed fields: action\_sequence, outcome. step\_index.jsonl allowed fields: observation, reasoning, action, result."\par
\noindent\hangindent=4em\hangafter=1\hspace*{4em}\},\par
\noindent\hangindent=4em\hangafter=1\hspace*{4em}"query": \{"type": "string", "description": "Case-insensitive regular expression matched against the selected evidence fields"\},\par
\noindent\hangindent=4em\hangafter=1\hspace*{4em}"fields": \{\par
\noindent\hangindent=5em\hangafter=1\hspace*{5em}"type": "array",\par
\noindent\hangindent=5em\hangafter=1\hspace*{5em}"items": \{\par
\noindent\hangindent=6em\hangafter=1\hspace*{6em}"type": "string",\par
\noindent\hangindent=6em\hangafter=1\hspace*{6em}"enum": ["action\_sequence", "outcome", "observation", "reasoning", "action", "result"]\par
\noindent\hangindent=5em\hangafter=1\hspace*{5em}\},\par
\noindent\hangindent=5em\hangafter=1\hspace*{5em}"description": "One or more exact fields allowed for the chosen index; do not mix fields from different indexes",\par
\noindent\hangindent=5em\hangafter=1\hspace*{5em}"minItems": 1\par
\noindent\hangindent=4em\hangafter=1\hspace*{4em}\},\par
\noindent\hangindent=4em\hangafter=1\hspace*{4em}"limit": \{"type": "integer", "description": "Maximum number of complete rows to return"\}\par
\noindent\hangindent=3em\hangafter=1\hspace*{3em}\},\par
\noindent\hangindent=3em\hangafter=1\hspace*{3em}"required": ["reason", "index", "query", "fields", "limit"]\par
\noindent\hangindent=2em\hangafter=1\hspace*{2em}\}\par
\noindent\hangindent=1em\hangafter=1\hspace*{1em}\}\par
\noindent\hangindent=0em\hangafter=1\hspace*{0em}\}\par
\endgroup
\end{tcolorbox}
\endgroup
\clearpage
\endgroup
\clearpage
\begingroup
\raggedbottom
\tcbset{evcuratorprompt/.style={
  enhanced, breakable, width=\linewidth,
  colback=black!2, colframe=black!50,
  colbacktitle=black!65, coltitle=white,
  fonttitle=\normalsize\bfseries, fontupper=\footnotesize\raggedright,
  boxrule=0.5pt, arc=2pt,
  left=8pt, right=8pt, top=6pt, bottom=6pt,
  toptitle=4pt, bottomtitle=4pt,
  before skip=0pt, after skip=0pt,
  before upper={\setlength{\parindent}{0pt}\setlength{\parskip}{0pt}}
}}

\par
\begingroup
\captionsetup{position=top,skip=0pt}
\captionof{table}{Native function schema for \texttt{inspect\_traj}. Only \texttt{reason} and \texttt{path} are required.}
\label{tab:inspect_traj_tool_schema}
\par\nopagebreak[4]\vspace{12pt}
\begin{tcolorbox}[evcuratorprompt,breakable=false,title={Navigation tool: \texttt{inspect\_traj} schema}]
\begingroup
\ttfamily\footnotesize\raggedright
\setlength{\parindent}{0pt}
\setlength{\parskip}{0pt}
\noindent\hangindent=0em\hangafter=1\hspace*{0em}\{\par
\noindent\hangindent=1em\hangafter=1\hspace*{1em}"type": "function",\par
\noindent\hangindent=1em\hangafter=1\hspace*{1em}"function": \{\par
\noindent\hangindent=2em\hangafter=1\hspace*{2em}"name": "inspect\_traj",\par
\noindent\hangindent=2em\hangafter=1\hspace*{2em}"description": "Read one exact trajectory\_path returned by corpus evidence to inspect the complete source episode. Step rows are already returned in full. Do not guess a path.",\par
\noindent\hangindent=2em\hangafter=1\hspace*{2em}"parameters": \{\par
\noindent\hangindent=3em\hangafter=1\hspace*{3em}"type": "object",\par
\noindent\hangindent=3em\hangafter=1\hspace*{3em}"properties": \{\par
\noindent\hangindent=4em\hangafter=1\hspace*{4em}"reason": \{"type": "string", "description": "Why this evidence path should answer the active question"\},\par
\noindent\hangindent=4em\hangafter=1\hspace*{4em}"path": \{"type": "string"\},\par
\noindent\hangindent=4em\hangafter=1\hspace*{4em}"offset": \{"type": "number"\},\par
\noindent\hangindent=4em\hangafter=1\hspace*{4em}"limit": \{"type": "number"\}\par
\noindent\hangindent=3em\hangafter=1\hspace*{3em}\},\par
\noindent\hangindent=3em\hangafter=1\hspace*{3em}"required": ["reason", "path"]\par
\noindent\hangindent=2em\hangafter=1\hspace*{2em}\}\par
\noindent\hangindent=1em\hangafter=1\hspace*{1em}\}\par
\noindent\hangindent=0em\hangafter=1\hspace*{0em}\}\par
\endgroup
\end{tcolorbox}
\endgroup
\clearpage
\endgroup
\clearpage
\begingroup
\raggedbottom
\tcbset{evcuratorprompt/.style={
  enhanced, breakable, width=\linewidth,
  colback=black!2, colframe=black!50,
  colbacktitle=black!65, coltitle=white,
  fonttitle=\normalsize\bfseries, fontupper=\footnotesize\raggedright,
  boxrule=0.5pt, arc=2pt,
  left=8pt, right=8pt, top=6pt, bottom=6pt,
  toptitle=4pt, bottomtitle=4pt,
  before skip=0pt, after skip=0pt,
  before upper={\setlength{\parindent}{0pt}\setlength{\parskip}{0pt}}
}}

\par
\begingroup
\captionsetup{position=top,skip=0pt}
\captionof{table}{System prompt for Navigation State Update.}
\label{tab:navigation_state_update_system}
\par\nopagebreak[4]\vspace{12pt}
\begin{tcolorbox}[evcuratorprompt,breakable=false,title={Navigation State Update: system message},fontupper={\fontsize{7.3}{8.2}\selectfont\raggedright}]
\textbf{Role}

\smallskip
Maintain the role-separated Navigation State for an outcome-aware task-time corpus investigation. You receive the current target context, current Navigation State, retrieval question, and one new corpus observation.

\smallskip
The State is the knowledge that a later skill may responsibly use for the current target. It is not a record of a source episode. Interpret the source evidence internally, then project only what transfers to the current target.

\smallskip
\textbf{Update procedure}

\smallskip
\textbf{1. Check relevance}

\smallskip
Decide whether the new observation actually answers the retrieval question. The retrieval question is only the investigation focus; it is not evidence.

\smallskip
If the observation does not answer the question, set evidence\_relevant=false and leave the complete Navigation State unchanged. Do not manufacture a conclusion merely because a retrieval returned text.

\smallskip
\textbf{2. Establish what happened in the source}

\smallskip
Use each evidence type only at the level it supports:

\smallskip
\begin{itemize}[leftmargin=*,nosep,topsep=1pt]
\item step\_index and chronological trajectory evidence may support exact commands, local transitions, and their ordering.
\item A failed attempt may support a failure lesson. It does not prove that the attempted action is a successful procedure.
\end{itemize}

\smallskip
Record what the evidence directly shows in source\_observation. Do not yet treat that observation as knowledge about the current instance.

\smallskip
\textbf{3. Project the source finding to the current target}

\smallskip
For each candidate lesson, ask whether it would still be valid if the source episode's identifiers, locations, and other instance-specific details were different while the current target stayed the same.

\smallskip
\begin{itemize}[leftmargin=*,nosep,topsep=1pt]
\item If it would remain valid, write that invariant or conditional lesson in transferable\_rule.
\item If it depends on a source-specific detail, do not copy that detail into transferable\_rule. Either express the lesson using information to be checked in the current observation, or do not retain it as target-usable knowledge.
\item A specific detail may appear in transferable\_rule only when the target or current observation explicitly confirms it for the current instance.
\item Put facts that can only be learned during current execution in target\_unknown. They are not questions for further corpus search.
\end{itemize}

\smallskip
The target and current observation are the only authorities for current-instance facts. Source repetition or source success does not make an identifier, location, or local state true in the current instance.

\smallskip
\textbf{4. Assign the projected lesson one role}

\smallskip
\begin{itemize}[leftmargin=*,nosep,topsep=1pt]
\item confirmed\_procedures: a target-applicable operation or causal relation whose procedural relation is established by the evidence;
\item search\_priors: a target-applicable candidate to check against current observation, stated without assuming a source-instance value;
\item failure\_lessons: a target-applicable failure cause, guard, or recovery;
\item open\_questions: uncertainties still worth another corpus investigation.
\end{itemize}

\smallskip
Confirmed means that the procedural relation is supported. It never means that a source episode's concrete value is confirmed for the current instance.

\smallskip
\textbf{5. Maintain the complete State}

\smallskip
Remove an open question when the updated State no longer needs it. Retain it only when a materially different corpus retrieval could answer it. Add a new question only when the evidence exposes such a procedural gap. Do not create or retain an open question whose answer depends on future observation of the current instance.

\smallskip
Because you return the complete State, also repair or remove an existing item when it contains a source-specific claim that the target or current observation does not support. Do not preserve a bad projection merely because it already has an id.

\smallskip
\textbf{6. Check consistency}

\smallskip
Before returning, compare every transferable\_rule with its target\_unknown. A rule must not assert a concrete value or condition that the same item says is unknown for the current target. Rewrite or remove any such item.

\smallskip
\textbf{Output contract}

\smallskip
Return the complete updated State through update\_search\_state. Do not store, write, or reproduce a draft skill.md or a source-episode summary. Call update\_search\_state exactly once and emit no separate answer.
\end{tcolorbox}
\endgroup
\clearpage
\endgroup
\clearpage
\begingroup
\raggedbottom
\tcbset{evcuratorprompt/.style={
  enhanced, breakable, width=\linewidth,
  colback=black!2, colframe=black!50,
  colbacktitle=black!65, coltitle=white,
  fonttitle=\normalsize\bfseries, fontupper=\footnotesize\raggedright,
  boxrule=0.5pt, arc=2pt,
  left=8pt, right=8pt, top=6pt, bottom=6pt,
  toptitle=4pt, bottomtitle=4pt,
  before skip=0pt, after skip=0pt,
  before upper={\setlength{\parindent}{0pt}\setlength{\parskip}{0pt}}
}}

\par
\begingroup
\captionsetup{position=top,skip=0pt}
\captionof{table}{User-message template for Navigation State Update.}
\label{tab:navigation_state_update_user}
\par\nopagebreak[4]\vspace{12pt}
\begin{tcolorbox}[evcuratorprompt,breakable=false,title={Navigation State Update: user message}]
\textbf{Navigation State Update}

\smallskip
\textbf{Inputs}

\smallskip
\textbf{Current target context}\par
\smallskip
\textbf{Target:}\par
\texttt{\{target\_task\}}

\smallskip
\textbf{Current observation:}\par
\texttt{\{current\_observation\}}

\smallskip
\textbf{Admissible commands:}\par
\texttt{\{admissible\_commands\}}

\smallskip
\textbf{Current Navigation State}\par
\smallskip
\texttt{\{current\_state\}}

\smallskip
\textbf{Retrieval question}\par
\smallskip
\texttt{\{retrieval\_question\_and\_reason\}}

\smallskip
\textbf{New evidence type}\par
\smallskip
\texttt{\{evidence\_kind\}}

\smallskip
\textbf{New corpus evidence}\par
\smallskip
\texttt{\{navigation\_observation\}}

\smallskip
\textbf{Required update}

\smallskip
\begin{enumerate}[leftmargin=*,nosep,topsep=1pt]
\item Set \texttt{evidence\_relevant} first.
\item If it is false, return the complete unchanged Navigation State with every \texttt{evidence\_used=false}.
\item If it is true, return complete replacements for \texttt{confirmed\_procedures}, \texttt{search\_priors}, \texttt{failure\_lessons}, and \texttt{open\_questions}.
\item Set \texttt{evidence\_used=true} only for entries created or changed by this evidence.
\end{enumerate}

\smallskip
Classify what the evidence supports; do not force it into an executable rule. For every knowledge entry, use \texttt{source\_observation} only for what happened in the source, use \texttt{transferable\_rule} only for the target-projected lesson, and use \texttt{target\_unknown} for facts that current execution must resolve. Ensure the rule does not assert anything declared unknown. Repair unsupported projections already present in the complete State instead of carrying them forward. An answered question may be removed, and a new question may be added when the evidence exposes a procedural uncertainty that another corpus retrieval can answer. Do not use \texttt{open\_questions} for hidden current-instance facts. The target-derived Goal is maintained outside this update and cannot be rewritten here.
\end{tcolorbox}
\endgroup
\clearpage
\endgroup
\clearpage
\begingroup
\raggedbottom
\tcbset{evcuratorprompt/.style={
  enhanced, breakable, width=\linewidth,
  colback=black!2, colframe=black!50,
  colbacktitle=black!65, coltitle=white,
  fonttitle=\normalsize\bfseries, fontupper=\footnotesize\raggedright,
  boxrule=0.5pt, arc=2pt,
  left=8pt, right=8pt, top=6pt, bottom=6pt,
  toptitle=4pt, bottomtitle=4pt,
  before skip=0pt, after skip=0pt,
  before upper={\setlength{\parindent}{0pt}\setlength{\parskip}{0pt}}
}}

\par
\begingroup
\captionsetup{position=top,skip=0pt}
\captionof{table}{System prompt and user-message template for Final Skill Synthesis.}
\label{tab:final_skill_synthesis_prompt}
\par\nopagebreak[4]\vspace{12pt}
\begin{tcolorbox}[evcuratorprompt,breakable=false,title={Final Skill Synthesis: messages}]
\textbf{System message}
\smallskip

You write one final task-time skill.md from the current target context and its role-separated Navigation State. The target defines what must be achieved; the State distinguishes what corpus evidence established from what it only suggested or warned against. Return only the complete \# skill.md.

\par\medskip\hrule\medskip
\textbf{User message}
\smallskip

\textbf{Final Skill Synthesis}

\smallskip
\textbf{Current target context}\par
\smallskip
\textbf{Target:}\par
\texttt{\{target\_task\}}

\smallskip
\textbf{Current observation:}\par
\texttt{\{current\_observation\}}

\smallskip
\textbf{Admissible commands:}\par
\texttt{\{admissible\_commands\}}

\smallskip
\textbf{Final Navigation State}\par
\smallskip
\texttt{\{final\_state\}}

\smallskip
\textbf{Synthesis rules}\par
\smallskip
\begin{itemize}[leftmargin=*,nosep,topsep=1pt]
\item Preserve the Goal and every target requirement.
\item Use confirmed\_procedures as the grounded procedural core, adapting their scope to the current target rather than copying a source episode.
\item Use search\_priors only as candidates for the executor to check against live observation and feedback. Do not present them as current facts.
\item Turn relevant failure\_lessons into concise guards or recovery guidance.
\item Open questions mark what the corpus did not establish. Keep those choices conditional on runtime evidence rather than inventing answers.
\item Specific details are allowed when the target or live context supports them; otherwise retain their uncertainty and function.
\item Produce concise end-to-end guidance, not an evidence report, search history, or restatement of the State.
\end{itemize}

\smallskip
Use this shape, omitting Checks and Recovery when it adds no useful information:
\smallskip
\begingroup
\ttfamily\footnotesize\raggedright
\setlength{\parindent}{0pt}
\setlength{\parskip}{0pt}
\noindent\hangindent=0em\hangafter=1\hspace*{0em}\# skill.md\par
\par\vspace{0.5\baselineskip}
\noindent\hangindent=0em\hangafter=1\hspace*{0em}\#\# Procedure\par
\noindent\hangindent=0em\hangafter=1\hspace*{0em}- Actionable guidance for the current target.\par
\par\vspace{0.5\baselineskip}
\noindent\hangindent=0em\hangafter=1\hspace*{0em}\#\# Checks and Recovery\par
\noindent\hangindent=0em\hangafter=1\hspace*{0em}- Include this section only when it adds useful guidance. Otherwise omit it.\par
\endgroup
\smallskip
Return only the complete \# skill.md.
\end{tcolorbox}
\endgroup
\clearpage
\endgroup

\end{document}